\documentclass[lettersize,journal]{IEEEtran}
\usepackage{array}
\usepackage[caption=false,font=normalsize,labelfont=sf,textfont=sf]{subfig}
\usepackage{stfloats}
\usepackage{url}
\usepackage{amsmath,amssymb,amsfonts}
\usepackage{algorithmic}
\usepackage{graphicx}
\usepackage{textcomp}

\usepackage[utf8]{inputenc}
\usepackage[T1]{fontenc}
\usepackage{verbatim}
\usepackage{bm}
\usepackage{caption2}
\usepackage{algorithm}
\usepackage{multirow}
\usepackage{hyperref}
\usepackage{float}
\usepackage{flushend}
\usepackage{makecell}
\usepackage{booktabs}

\usepackage{pifont}
\newcommand{\cmark}{\ding{51}}%
\newcommand{\xmark}{\ding{55}}%
\usepackage{soul}
\usepackage{tikz}

\def\BibTeX{{\rm B\kern-.05em{\sc i\kern-.025em b}\kern-.08em
    T\kern-.1667em\lower.7ex\hbox{E}\kern-.125emX}}
\usepackage{balance}
\begin{document}

\title{ Constrained Reinforcement Learning using Distributional Representation for Trustworthy Quadrotor UAV Tracking Control}
\author{Yanran Wang, David Boyle
\thanks{This work was partially supported by NSF-UKRI [grant number NE/T011467/1]; and the Engineering and Physical Sciences Research Council [grant number EP/X040518/1].}}

\markboth{}%
{How to Use the IEEEtran \LaTeX \ Templates}

\maketitle

\begin{abstract}
Simultaneously accurate and reliable tracking control for quadrotors in complex dynamic environments is challenging. {The chaotic nature of} aerodynamics, derived from drag forces and moment variations, {makes precise identification} difficult. {Consequently}, {many existing} quadrotor tracking systems treat {these aerodynamic effects} as simple `disturbances' in conventional control approaches. We propose a novel {and} interpretable trajectory tracker integrating a {d}istributional Reinforcement Learning {(RL)} disturbance estimator for unknown aerodynamic effects with a Stochastic Model Predictive Controller (SMPC). {Specifically, t}he proposed estimator `Constrained Distributional R{E}inforced{-D}isturbance{-}estimator' (ConsDRED) {effectively} identifies uncertainties between {the} true and estimated values of aerodynamic effects. Control parameterization employs {s}implified {a}ffine {d}isturbance {f}eedback to {ensure} convexity, which is seamlessly integrated with {the} SMPC. We theoretically guarantee that ConsDRED achieves an optimal global convergence rate, and sublinear rates if constraints are violated with certain error decreases as neural network dimensions increase. To demonstrate practicality, we show convergent training{,} in simulation and real-world experiments, and empirically verify that ConsDRED is less sensitive to hyperparameter settings compared with canonical constrained RL. {O}ur system {substantially} improves accumulative tracking errors by at least 70\%{,} compared with the recent art. Importantly, the proposed ConsDRED-SMPC {framework} balances the trade-off between pursuing high performance and obeying conservative constraints for practical implementations\footnote{Our code and all research artefacts will be made openly available upon acceptance of our manuscript for publication. Video figures in support are also available: \url{https://github.com/Alex-yanranwang/ConsDRED-SMPC}.}.
\end{abstract}


\section{Introduction}
\label{sec:introduction}
\IEEEPARstart{A}{ccurate} trajectory tracking for autonomous Unmanned Aerial Vehicles (UAVs), such as quadrotors, is necessary for maintaining autonomy. Although industrial applications of autonomous UAVs \cite{mishra2020drone,dissanayaka2023review} have attracted much attention in recent years, precisely and reliably tracking high-speed and high-acceleration UAV trajectories is an extremely challenging control problem, particularly in unknown environments with unpredictable aerodynamic forces \cite{wei2023safe,xing2023active}.

To achieve {a} safe, precise and reliable quadrotor trajectory tracking, two main problems need to be solved: How to achieve accurate and trustworthy estimation (or modelling) of the variable aerodynamic effects using limited onboard computational resources? {Then h}ow can the whole control framework be integrated with the aerodynamic effect estimation to track trajectory references precisely and reliably?

Previous work has shown that the primary source of quadrotor uncertainties are aerodynamic effects deriving from drag forces and moment variations caused by the rotors and the fuselage \cite{torrente2021data}. Wind tunnel experiments show that aggressive maneuvers at high speed, e.g., greater than 5 $ms^{-1}$, introduce significant positional and attitude tracking errors \cite{faessler2017differential}. Although modelling aerodynamic effects precisely in static environments like no-gust indoor areas, data-driven approaches - e.g., Gaussian Processes (GP) \cite{torrente2021data,wang2022kinojgm} and neural networks \cite{spielberg2021neural,dong2022deep} - perform poorly in complex environments. The main reason is the generalization ability: their training datasets, collected from simulated platforms or real-world historical records, do not fully describe the complex environments.

Reinforcement Learning (RL) can solve the complex and changeable sequential decision-making process \cite{zhang2021model} with iteratively interactive learning. The key challenge of most existing RL approaches \cite{lillicrap2015continuous} is: policy optimization relies on nebulous black/grey-boxes where deep neural networks are poorly explainable, and the global convergence is historically unstable \cite{beggs2005convergence, garcia2015comprehensive, xu2021crpo}. These challenges mean that traditional RL results are often irreproducible \cite{lynnerup2020survey} - i.e., the same hyperparameter setting yields a considerable discrepancy between the stated and reproduced results. Because of this uncontrollable and non-interpretable convergence process, industrial applications of RL are often unacceptable, particularly in safety-critical applications such as autonomous robots \cite{wang2022kinojgm,nguyen2021model} and financial services \cite{ostmann2021ai}.

To address the stated issues, we propose \textbf{Cons}trained \textbf{D}istributional \textbf{Re}inforced-\textbf{D}isturbance-estimation for Stochastic MPC (ConsDRED-SMPC), a systematic, trustworthy and feasible quadrotor trajectory tracking framework for use under high variance aerodynamic effects. { To our best knowledge, this study represents the first attempt within the robot learning community to develop a trustworthy interpretation of an RL-based disturbance estimator through practical implementation.} The details are as follows:
\begin{enumerate}
    \item Aerodynamic Disturbance Estimator: a Constrained Distributional R{E}inforced-{D}isturbance-estimator (ConsDRED; Algorithm~\autoref{DRL_ConsDRED}), is proposed for aerodynamic disturbance estimation. This estimation relies on wind estimation obtained from VID-Fusion \cite{ding2020vid} (see Fig.~\ref{RL_Control_framework} for more details).
    ConsDRED builds upon prior QR-DQN \cite{dabney2018distributional} and QUOTA \cite{zhang2019quota} insofar as ConsDRED is a quantile-approximated constrained distributional RL which uses a set of quantiles to approximate the full value distribution in a Constrained RL (CRL, also known as safe RL) framework. Section \uppercase\expandafter{\romannumeral5}-A shows that ConsDRED can not only guarantee the convergence (i.e., \textit{Theorem 1}), but also achieve at least an $\Theta(1/{\sqrt{K}})$ convergence rate (\textit{Proposition 4}) to local policy evaluation (i.e., the distributional temporal difference (TD) learning defined in \autoref{TD_learning}) and at least an $\Theta(1/{\sqrt{T}})$ convergence rate (\textit{Theorem 2}) to global policy optimization, respectively.
    
    \item Trajectory Tracker: Similar to \cite{zhang2021stochastic}, a Simplified Affine Disturbance Feedback (SADF) is used for control parameterization in SMPC (Algorithm~\autoref{SADF_SMPC}), where the convexity can be guaranteed in this process \cite{lofberg2003approximations} and computational complexity can be reduced. Unlike prior work assuming zero mean disturbance, we consider the control performance and stability under non-zero-mean disturbance. We use an Input-to-State Stability (ISS) \cite{jiang2001input} property to find conditions that imply stability and convergence of the tracker. 
    \item The ConsDRED-SMPC framework is proposed to track quadrotor trajectory accurately under high variance aerodynamic effects. The overall control framework is shown in Fig.~\ref{RL_Control_framework}. In Section \uppercase\expandafter{\romannumeral5}-B, the closed-loop stability of ConsDRED-SMPC is demonstrated under the Lipschitz Lyapunov function \cite{rifford2000existence}.
\end{enumerate}

Our contributions can be summarized as follows:
\begin{itemize}
\item[1)]
ConsDRED, a constrained distributional RL {employing} quantile approximation, {demonstrates significant capability in} adaptively estimating variable aerodynamic disturbances. In the cases tested, we show that ConsDRED outperforms the state-of-the-art CRL, such as CRPO \cite{xu2021crpo}, and prior {d}istributional RL approaches, such as QuaDRED \cite{wang2022interpretable}. More importantly, ConsDRED achieves at least an $\Theta(1/{\sqrt{T}})$ global convergence rate, which offers an interpretable and trustworthy guarantee to the practical training implementation.
\end{itemize}
\begin{itemize}
\item[2)]
{ConsDRED-SMPC,} a {quadrotor} trajectory tracking framework, integrates ConsDRED, i.e., the aerodynamic disturbance estimator, into a stochastic optimal control problem.
\end{itemize}
\begin{itemize}
\item[3)]
Convergence and stability guarantees: mathematical proofs are provided for the convergence rate of the constrained distributional-RL-based estimator {, i.e., ConsDRED,} and the closed-loop stability of the stochastic-MPC-based tracker{,} with consideration of non-zero-mean and bounded disturbances.
\end{itemize}

\begin{table*}[t]
    \centering
    \caption{Comparison of Tracking Control Methods for Quadrotor UAVs}
    \label{method_comparison}
    \setlength{\tabcolsep}{0.85mm}{
    \begin{tabular}{c c c c c c c}
    \toprule
    \multirow{3}{*}{\textbf{Methods}} & \multirow{3}{*}{\textbf{Algorithm}} & \multicolumn{5}{c}{\textbf{Safe and trustworthy consideration}}\\ 
    \cline{3-7}
    & & \textbf{\makecell[c]{Non-conservative}} & \textbf{\makecell[c]{Unknown structure \\ uncertainties}}  & \textbf{Safety constraints} & \textbf{\makecell[c]{Combining with \\model-based control}} & \textbf{Formal stability}\\
    \hline
    \multirow{3}{*}{Robust control} & Abdelmoet et. al., 2016 \cite{abdelmoeti2016robust} & \xmark & \cmark & \cmark & \cmark & \cmark \\
    & Raffo et. al., 2016 \cite{raffo2016nonlinear}& \xmark & \cmark & \cmark & \cmark & \cmark \\
    & Nguyen et. al., 2021 \cite{nguyen2021model} & \xmark & \cmark & \cmark & \cmark & \xmark \\[0.18cm]
    \multirow{2}{*}{Adaptive control} & Zhang et. al., 2018 \cite{zhang2018aerodynamic} & \cmark & \xmark & \cmark & \cmark & \cmark \\
    & Zhang et. al., 2021 \cite{zhang2021model}& \cmark & \xmark & \cmark & \cmark & \cmark \\[0.18cm]
    \makecell[c]{Adaptive dynamic\\ programming} & Dou et. al., 2021 \cite{dou2021robust} & \cmark & \cmark & \xmark & \xmark & \cmark \\[0.18cm]
    \multirow{3}{*}{Data-driven approach} & Torrente et. al., 2021 \cite{torrente2021data} & \cmark & \xmark & \xmark & \cmark & \xmark \\
    & Wang et. al., 2022 \cite{wang2022kinojgm}& \cmark & \cmark & \xmark & \cmark & \xmark \\
    & Spielberg et. al., 2021 \cite{spielberg2021neural}& \cmark & \cmark & \xmark & \cmark & \xmark \\[0.18cm]
    Our method &  & \cmark & \cmark & \cmark & \cmark & \cmark \\
    \bottomrule
    \end{tabular}}
\end{table*}

\section{Motivation and Related Work}
\subsection{Estimation of Aerodynamic Effects}
Aerodynamics explains how the air moves around things \cite{bertin2021aerodynamics}, where prominent aerodynamic disturbances appear at flight speeds of 5 $ms^{-1}$ in wind tunnel experiments \cite{faessler2017differential}. These effects acting on quadrotors are chaotic and hard to model directly, as they are generated from a combination of the individual propellers and airframe \cite{hoffmann2007quadrotor}, turbulent effects caused by rotor–rotor and airframe–rotor interactions \cite{russell2016wind}, and the propagation of other turbulence \cite{kaya2014aerodynamic}.

Most current approaches to quadrotor trajectory tracking treat aerodynamic effects as simple external disturbances and do not account for higher-order effects or attempt to deviate from a determined plan \cite{tal2020accurate,lee2019robust,bicego2020nonlinear}. While these solutions are efficient and feasible for lightweight on-board computers, aggressive flight at high speed (e.g., $\geq$ 5 $ms^{-1}$) introduce{\ colour {red}s} large tracking errors \cite{faessler2017differential}. Recent data-driven approaches, such as GPs \cite{torrente2021data,wang2022kinojgm} and neural networks \cite{spielberg2021neural} combined with Model Predictive Control (MPC), show accurate modelling of aerodynamic effects. However, the nonparametric nature of GP causes them {to} perform poorly in complex environments, leading to escalating computational demands as the training set size increases \cite{saviolo2023learning}, particularly when large datasets contain drastic changes in wind speed and heading. In these instances, learning-based (neural networks) approaches perform better than GP-based approaches \cite{loquercio2021learning,singh2021evolving,wu2023event,o2022neural}.  Although combining model-based estimation with data-driven learning appears more effective \cite{jia2023evolver}, it may not be suitable for certain variable disturbances such as payloads, pushes or collisions, due to the conservative nature of \textit{Assumption 1} in \cite{jia2023evolver}. Consequently, achieving adaptability and robustness in complex environments is still challenging, primarily because training datasets and models are static and most do not completely identify the complex environmental dynamics \cite{saviolo2023learning}.

\subsection{Reinforcement Learning in real physical systems: insights from distributional and Constrained perspectives}

\textit{Simulation to reality in reinforcement learning:} In real-world scenarios, the inherent challenges posed by observation dimensionality, task complexity, or exploration difficulty make the utilization of synthetic experience a nature choice for practical implementation of RL \cite{rizzardo2023sim}. Simulation to Reality (Sim-to-Real) RL methods \cite{zhu2023event, muratore2022robot, hofer2021sim2real} leverage simulation platforms to efficiently train policies within virtual replicas of the intended environment. Subsequently, these trained policies are transferred to the real-world domain. To overcome the Sim-to-Real gap, approaches proposed in \cite{torrente2021data, saviolo2022physics} enhance the accuracy of model identification, while domain randomization techniques, as employed in \cite{tiboni2023dropo, horvath2022object}, contribute to improving learning generalization.

\textit{Distributional representation in reinforcement learning:} In comparison to existing data-driven approaches, RL, an interactive learning process, can learn complex and changeable disturbances, i.e., the errors between true and estimated values, using much less model information \cite{zhang2021model}. However, to maximize the accumulated rewards, policy optimization biases toward actions with high variance value estimates, since some of these values will be overestimated by random chance \cite{ma2021conservative}. These actions should be avoided in risk-sensitive applications such as real-world autonomous navigation. Recent work on distributional RL \cite{bellemare2017distributional} proposes to approximate and parameterize the entire distribution of future rewards, instead of the expected value. Distributional RL algorithms have been shown to achieve promising results on continuous control domains \cite{hessel2018rainbow}. In principle, more complete and richer value-distribution information is provided to enable a more stable learning process \cite{bellemare2017distributional}. Previous distributional RL algorithms parameterize the policy value distribution in different ways, including canonical return atoms \cite{bellemare2017distributional}, the expectiles \cite{rowland2019statistics}, the moments \cite{nguyen2021distributional}, and the quantiles \cite{dabney2018distributional,zhang2019quota}. The quantile approach is especially suitable for autonomous UAV trajectory tracking \cite{wang2022interpretable,wang2023quadue} due to its risk-sensitive policy optimization.

\textit{Constrained reinforcement learning addressed through primal-dual approaches:} In the setting above, the agent is allowed to explore the entire state and action space without any constraints, where one issue is \cite{gu2022review,zhao2023stable}: \textbf{how can we guarantee safety when we apply RL for real-world applications}? In CRL, however, the space is explored under safe constraints, where policy optimization aims to maximize the rewards whilst satisfying certain safe constraints \cite{garcia2015comprehensive}. Two main categories of CRL solutions are the \textbf{primal} and \textbf{primal-dual} approaches. The primal-dual approaches, combining the value function with a sum of constraints weighted by corresponding Lagrange multipliers, are frequently adopted in CRL, e.g., the well-known RCPO \cite{tessler2018reward}, CPPO \cite{stooke2020responsive}, and PPO \cite{ding2021provably}. Theoretically, in these primal-dual methods, \cite{zhao2023stable,tessler2018reward} gives an asymptotic convergence guarantee and \cite{ding2021provably} achieves a regret bound for linear constrained MDP. However, the primal-dual approach can be severely sensitive to these {hyperparameters}: tuning the learning rates and threshold of the dual Lagrange multipliers. This means that the primal-dual approach is determined by the initialization of those hyperparameters, {thus acquiring} additional costs in hyperparameter tuning \cite{tessler2018reward}. On the other side, the primal approach \cite{chow2018lyapunov, liu2020ipo, xu2021crpo} designs the objective functions diversely without the dual Lagrange variables, {receiving} much less attention than the primal-dual approach. CRPO \cite{xu2021crpo} proposes the first primal algorithm with a provable convergence guarantee. However, it is only established on simpler $2$-layer neural networks, which is impractical for large-scale search space, i.e., extension to multi-layer neural networks.\vspace{-0.045cm}

To address the concerns above, the three perspectives of RL research have emerged. We use a distributional RL based on quantile approximation for modelling the state-action distribution, incorporating Sim-to-Real RL training (detailed setting provided in Section \uppercase\expandafter{\romannumeral6}-A). Then, we employ a primal algorithm for constrained policy optimization. Compared with \cite{wang2022interpretable}, our proposed ConsDRED guarantees $\Theta(1/{\sqrt{T}})$ convergence rate in both local policy evaluation and global policy optimization. Building on \cite{xu2021crpo}, our work extends the convergence guarantees to (multi) $H$-layer neural networks (see \textit{Proposition 4} and \textit{Theorem 2}) for practical implementation. Our numerical experiments also demonstrate that ConsDRED achieves at least $61\%$ and $11\%$ improvement of tracking errors than CRPO \cite{xu2021crpo} in the simulated and real-world scenarios, respectively (see Section \uppercase\expandafter{\romannumeral6}-B).

\subsection{Tracking Control Framework under Disturbances}
Quadrotor UAV tracking control for uncertain systems can be categorized into four main approaches: \textbf{1)} Robust control adopts a "worst-case" formulation to handle bounded uncertainties and disturbances \cite{wei2022online, zhou1998essentials}. However, in real-world scenarios with uncertainties, this worst-case design can lead to suboptimal and overly conservative control actions \cite{mayne2016robust, mesbah2016stochastic}. \textbf{2)} Adaptive control is capable of dealing with varying uncertainties with unknown boundaries, but it assumes that uncertainties are linearly parameterized with known structure and unknown parameters \cite{zhang2021model,zhang2018aerodynamic}. \textbf{3)} Adaptive Dynamic Programming (ADP) belongs to a specific class of RL algorithms that combine dynamic programming and RL principles. ADP emphasizes updating value functions or policies using dynamic programming principles, making it suitable for handling large state and action spaces. However, ADP also has some weaknesses in quadrotor control \cite{dou2021robust}, which typically rely on accurate models for effective convergence, and uncertainty in the model can impact the quality of the learned policy. \textbf{4)} Data-driven approaches, such as Gaussian Processes (GP) \cite{torrente2021data, wang2022kinojgm} and neural networks \cite{wei2023safe,spielberg2021neural}, efficiently model aerodynamic effects in static environments like indoor areas without gusts. However, these methods perform poorly in complex environments like forests due to their limited generalization capabilities. A concise comparison between existing methods and our proposed method is presented in \autoref{method_comparison}.

Robust MPC for tracking control of uncertain systems like quadrotors \cite{wu2023event,wei2022online,sun2022comparative,saviolo2023active} is rapidly developing thanks to advances in hardware and algorithm efficiency \cite{nguyen2021model}. To avoid the conservatism of the worst case design, Stochastic MPC (SMPC) \cite{schwarm1999chance,huanca2023design} uses the probabilistic descriptions, such as stochastic constraints (also called chance constraints), to predict probability distributions of system states within acceptable levels of risk in the receding-horizon optimization \cite{primbs2009stochastic}.

The core challenges for SMPC include: 1) optimizing the feedback control laws over arbitrary nonlinear functions \cite{munoz2020convergence}; 2) the chance constraints are non-convex and intractable \cite{mesbah2016stochastic,zhang2021stochastic}; and 3) the computational complexity will grow dramatically as more uncertainties are added. To address the first challenge, one solution is to use an affine parameterization of the control policy over finite horizons. However, this approach cannot guarantee convexity, i.e., the second challenge, where the policy set may still be convex \cite{lofberg2003approximations}. Another solution is an Affine Disturbance Feedback (ADF) control parameterization, proposed in \cite{goulart2008input}. This ADF control parameterization can address the first two challenges, optimizing the dynamic function and guaranteeing the decision variables to be convex. However, the main weakness is that the computational complexity grows quadratically with the prediction horizon, i.e., the third challenge. To overcome this difficulty, a Simplified Affine Disturbance Feedback (SADF) proposed in \cite{zhang2021stochastic}, where the SADF is equivalent to ADF but a finite-horizon optimization can be computed more efficiently using CasADi \cite{andersson2019casadi}, a nonlinear MPC solver. \cite{zhang2021stochastic} achieves good results by implementing SADF with zero-mean disturbance. However, how the SADF would perform on systems{, such as a quadrotor,} with non-zero-mean disturbances is unclear.

\section{Problem Formulation}
\subsection{Dynamic Model of Quadrotors}
A quadrotor dynamic model has six Degrees of Freedom (DoF), i.e., three linear motions and three angular motions \cite{torrente2021data}. Let $\bm{x} = [\bm{P}_{WB},\bm{V}_{WB},\bm{q}_{WB},\bm{\omega} _{B}]^T\in\mathbb{X}\subseteq\mathbb{R}^{n}$, $\bm{u}= [\bm{c},\bm{\tau}_B]^T$ and $\bm{e}_{f}$ be the state, control input and aerodynamic effect, respectively. $\bm {P}_{WB}$, $\bm{V}_{WB}$ and $\bm{q}_{WB}$ are the position, linear velocity and orientation of the quadrotor, expressed in the world frame. $\bm{\omega}_{B}$ is the angular velocity \cite{wang2022kinojgm} expressed in the body frame. $\bm{c}$ and $\bm{\tau}_B$ are the collective thrust, defined as $\bm{c} = [0,0,\sum T_i]^{\rm{T}}$, and the body torque, defined in (Equation 2 in) \cite{torrente2021data}, respectively. $T_i$ is the thrust of the $i$-th ($i\in[0,3]$) motor. We consider the continuous-time nominal model of quadrotors defined in \autoref{quadrotor_dynamics}:
\begin{equation}
\begin{aligned}
&\dot{\bm{P}}_{WB} = \bm{V}_{WB}\qquad\dot{\bm{V}}_{WB} = \bm{g}_W + \frac{1}{m}(\bm{q}_{WB}\odot \bm{c} + \bm{e}_f)\\
&\dot{\bm{q}}_{WB} = \frac{1}{2}\Lambda(\bm{\omega} _{B})\bm{q}_{WB}\quad\dot{\bm{\omega}} _{B} = \bm{J}^{-1}(\bm{\tau}_B-\bm{\omega}\times J\bm{\omega} _{B})
\end{aligned}
\label{quadrotor_dynamics}
\end{equation}
where $\bm{g}_W = [0,0,-\bf{g_w}]^{\rm{T}}$. The operator $\odot$ denotes a rotation of the vector by the quaternion. The skewsymmetric matrix $\Lambda(\bm{\omega})$ is defined in \cite{wang2022kinojgm}.

We discretize and linearize \cite{goulart2008input} \autoref{quadrotor_dynamics} for MPC over a finite horizon $N$:
\begin{equation}
\bm{x}_{t} = \bm{A}x_{0|t} + \bm{B}\bm{u}_{t} + \bm{G}\bm{w}_{t}
\label{linear_discrete_system}
\end{equation}
where $\bm{x}_{t}=[x_{0|t}^{\rm{T}}, x_{1|t}^{\rm{T}}, ..., x_{N|t}^{\rm{T}}]^{\rm{T}}$ and $\bm{u}_{t}=[u_{0|t}^{\rm{T}}, u_{1|t}^{\rm{T}}, ..., u_{N|t}^{\rm{T}}]^{\rm{T}}$ are the sequential discrete-time states and inputs of the quadrotor nominal model. $\bm{w}_{t}=[w_{0|t}^{\rm{T}}, w_{1|t}^{\rm{T}}, ..., w_{N|t}^{\rm{T}}]^{\rm{T}}$, a sequential stochastic disturbance (over a horizon of $N$) caused by aerodynamic effects, inaccuracy in VID-Fusion \cite{ding2020vid}, dynamic discrepancy of nominal model and linearization error, equals to the action output of ConsDRED (an aerodynamic disturbance estimator) proposed in Section \uppercase\expandafter{\romannumeral4}-A).  $\bm{A}$, $\bm{B}$ and $\bm{G}$ are matrices defined in \cite{zhang2021stochastic}.

\subsection{Constrained Distributional Markov Decision Process}
We consider a Constrained Distributional Markov Decision Process (CDMDP), combining a discounted Constrained Markov Decision Process (CMDP) \cite{altman1999constrained} and a distributional Bellman equation \cite{bellemare2023distributional}.

The CMDP is formulated as a discounted Markov Decision Process (MDP) with additional constrained objectives, i.e., a tuple $\left \langle S, A, P, R, g, \gamma \right \rangle$, where $S$ is a finite set of states ${ \left\{ \bm{s} \right\}}$, $A$ is a finite set of actions ${ \left\{ \bm{a} \right\}}$, $P: S\times A\rightarrow S$ is a finite set of transition probabilities ${ \left\{ p \right\}}$, $R: S\times A\times S \rightarrow \mathbb{R}$ is a finite set of bounded immediate rewards ${ \left\{ r \right\}}$, $g$ is a unity function where the agent is constrained into a `safe' state, and $\gamma\in[0,1]$ is the discount rate. A stationary policy $\pi$ maps one state $\bm{s}$ to one action $\bm{a}$. The following constrained problem is presented as a CMDP:
\begin{equation}
\underset{\pi}{\rm{max}}\: \mathcal{J}_r(\bm{\pi}), \quad {\rm{s.t.}}\quad \mathcal{J}^{i}_g(\bm{\pi})\leq \bm{b}_i,\quad i=1,...,p
\label{constrained_rl_problem}
\end{equation}
where $\mathcal{J}_r(\bm{\pi}):= {\mathbb{E}}[{\sum\limits_{t=0}^{\infty}\gamma^{t}}r(\bm{s}_t,\bm{a}_t)|\pi, {\bm{s}_0}=s]$ and $\mathcal{J}^{i}_g(\bm{\pi}):= {\mathbb{E}}[{\sum\limits_{t=0}^{\infty}\gamma^{t}}g^{i}(\bm{s}_t,\bm{a}_t)|\pi, {\bm{s}_0}=s]$ are the value function associated with the reward $r$ and the utility $g$, respectively, and where $\bm{s}_{t+1}\sim p(\cdot|\bm{s}_t,\bm{a}_t)$ and $\bm{a}_{t}\sim\pi(\cdot|\bm{s}_t)$ at each time step, $t$.

The distributional Bellman equation \cite{bellemare2017distributional}, the aim of which is different from traditional RL, maximizes the expectation of value-action function $Q$. In the policy evaluation setting, given a deterministic policy $\pi$, the state-action distribution $Z^\pi$ and the \textit{Bellman operator} $\mathcal{T}^\pi$ are defined as \cite{bellemare2017distributional,dabney2018distributional}:
\begin{equation}
\mathcal{T}^{\pi} Z(\bm{s},\bm{a})\overset{D}{:=} R(\bm{s},\bm{a})+\gamma Z(S',A')
\label{Bellman_Pred_D_RL}
\end{equation}
where the state $\bm{s}=[\bm{P}_{WB},\bm{V}_{WB},\bm{q}_{WB},\bm{\omega} _{B},\bm{e}_{f}]^T \in S$, the action $\bm{a}=\bm{w} \in A$, and $S'$ and $A'$ are the sets of states and actions sampled at the next time step.

The following assumptions are made: 

\textit{Assumption 1}: Matrix G is column full rank.

\textit{Assumption 2}: The aerodynamic effect $\bm{e_{fk}}$ is available with no delay at each sampling timestamp.

\textit{Proposition 1}: There exists a control law $\bm{u}_b$ that ensures the nominal model $\bm{f}(\bm{x}_{k},\bm{u}_{k}, \bm{e_{fk}})$ is ISS if the stochastic disturbance $\bm{w}_t$ is an independent and identically distributed (i.i.d.) zero-mean distribution, i.e., $\mathbb{E}{(w_k)}=0$.

\textit{Proof}: Based on \textit{Assumption 2}, the nominal model $\bm{f}(\bm{x}_{k},\bm{u}_{k}, \bm{e_{fk}})$ is seen as $\bm{f}(\bm{x}_{k},\bm{u}_{k}^{'})$, in which the aerodynamic force $\bm{e_{fk}}$ is a constant term. Then, based on \cite{jiang2001input,munoz2020convergence,goulart2008input}, $\bm{f}(\bm{x}_{k},\bm{u}_{k}^{'})$ is an ISS-Lyapunov function. \hfill $\blacksquare$

\begin{figure}[t]
  \centering
  \includegraphics[scale=0.54]{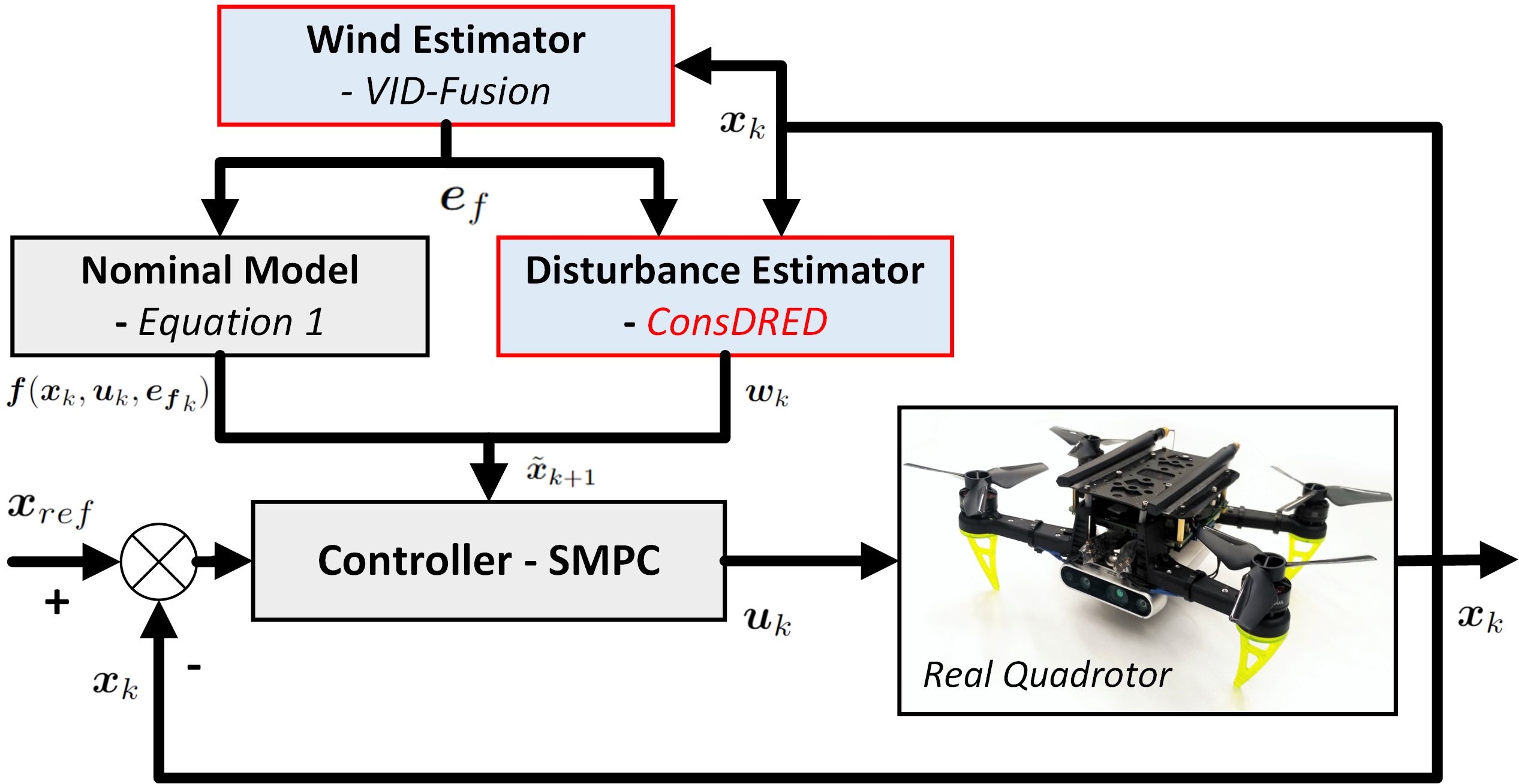}
  \caption{ConsDRED-SMPC: (i) a wind estimator, i.e., VID-Fusion \cite{ding2020vid}; (ii) an aerodynamic disturbance estimator, i.e., the proposed ConsDRED (described by Algorithm~\autoref{DRL_ConsDRED}); and (iii) a controller, i.e., SMPC.}
  \label{RL_Control_framework}
\end{figure}

\section{ConsDRED-SMPC}
In this section, we present the proposed ConsDRED-SMPC control framework. Traditional non-interacted methods, e.g., GP \cite{torrente2021data} and RDRv \cite{faessler2017differential}, are insufficient for quadrotor dynamic disturbance estimation. This work addresses the limitation and proposes a novel and feasible disturbance estimation with continuous environmental interactions for variable winds.

\subsection{Constrained Distributional Reinforced Estimation \label{D_RL}}
\noindent \textbf{Policy Evaluation}: to perform the policy evaluation in \autoref{Bellman_Pred_D_RL} and Algorithm~{\autoref{DRL_ConsDRED}} (Line 5), we guarantee the contraction of the \textit{Bellman operator} $\mathcal{T}^\pi$ over the Wasserstein Metric (see \autoref{contraction_WM}) and then adopt Wasserstein distance to obtain the distance between the target $\mathcal{T}^{\pi} Z$ and the prediction $Z$.

The Wasserstein Metric, also known as the Mallows metric, is a true probability metric with no disjoint support issues. A contraction is proved in \cite{dabney2018distributional} over the Wasserstein Metric:
\begin{equation}
\overset{-}{d}_{\infty}(\Pi_{W_1}\mathcal{T}^{\pi}Z_1,\Pi_{W_1}\mathcal{T}^{\pi}Z_2)\leq \overset{-}{d}_{\infty}(Z_1,Z_2)
\label{contraction_WM}
\end{equation}
where $W_p$, $p\in[1,\infty]$ denotes the $p$-Wasserstein distance. $\overset{-}{d}_{p}:={\rm{sup}}W_p(Z_1,Z_2)$ denotes the maximal form of the  $p$-Wasserstein metrics. $T$, as defined in \autoref{Bellman_Control_D_RL} below, is a distributional Bellman optimality operator, and $\Pi_{W_1}$ is a quantile approximation under the minimal 1-Wasserstein distance $W_1$.

Based on the above contraction guarantees, we employ TD learning to estimate the state-action value distribution $Z$, where in each iteration, we propose:
\begin{equation}
\begin{aligned}
&\zeta^{i}_{k+1}(\bm{s},\bm{a})=\zeta^{i}_{k}(\bm{s},\bm{a})+l_{td}\bm{\Delta}^{i}_{k}=\zeta^{i}_{k}(\bm{s},\bm{a})+l_{td}\times\\
&\overset{-}{d}_{\infty}(\Pi_{W_1}\mathcal{T}^{\pi}(h_i(\bm{s},\bm{a},\bm{s}')+\gamma\zeta^{i}_{k}(\bm{s}')),\Pi_{W_1}\mathcal{T}^{\pi}\zeta^{i}_{k}(\bm{s},\bm{a}))
\label{TD_learning}
\end{aligned}
\end{equation}
where $\zeta^{i}_k\in S\times A$ is the estimated distribution of the state-action distribution $Z$ in the $k$-th TD-learning-iteration for all $i=0,...,p$, and $l_{td}$ is the TD learning rate. $h_i: S\times A\times S\rightarrow \mathbb{R}$ maps $(\bm{s},\bm{a},\bm{s}')$ to a \textit{Real Number}, of which the definition is $h_i=r$ when $i=0;$ and $h_i=g^i$ when $i\in[1,p]$. The distributional TD error $\bm{\Delta}^{i}_{k}$ in \autoref{TD_learning} is calculated by $\overset{-}{d}_{\infty}(\Pi_{W_1}\mathcal{T}^{\pi}(h_i(\bm{s},\bm{a},\bm{s}')+\gamma\zeta^{i}_{k}(\bm{s}')),\Pi_{W_1}\mathcal{T}^{\pi}\zeta^{i}_{k}(\bm{s},\bm{a}))$.

It has been shown in \cite{dabney2018distributional}, \cite{tang2022nature} that the quantile approximation of the Bellman operator $\Pi_{W_1}\mathcal{T}$ is a contraction over the maximal form of the
Wasserstein distance $\overset{-}{d}_{\infty}$. Therefore, after performing $K_{td}$ iterations, $\zeta^{i}_k$ will be contracted and converge to a fixed point $\zeta^{i}_{*}\in S\times A$, and then we let $\overset{-}{Z^{i}_{t}}(\bm{s},\bm{a})=\zeta^{i}_{K_{td}}(\bm{s},\bm{a}),\forall(\bm{s},\bm{a})\in S\times A \ and\ \forall i\in[0,p]$.

\noindent \textbf{Constraint Estimation}: the constraint function $\mathcal{J}^{i}_g(\bm{\pi})$ is approximated via weighted utility function $g^{i}(\bm{s}_t,\bm{a}_t)$, i.e., $\overset{-}{\mathcal{J}^{i}_g}(\bm{\pi})={\mathbb{E}}[{\sum\limits_{t=0}^{\infty}\gamma^{t}}g^{i}(\bm{s}_t,\bm{a}_t)|\pi, {\bm{s}_0}=s]={\mathbb{E}}[\zeta^{i}_{K_{td}}(\bm{s},\bm{a})],\ \forall i\in[1,p]$. Then we have $|\overset{-}{\mathcal{J}^{i}_{g}}(\bm{\pi})-\mathcal{J}^{i}_g(\bm{\pi})|=|\mathbb{E}[\zeta^{i}_{K_{td}}(\bm{s},\bm{a})]-\mathbb{E}[g^i(\bm{s},\bm{a})]|\leq \left \| \zeta^{i}_{K_{td}}(\bm{s},\bm{a})-g^i(\bm{s},\bm{a}) \right \|,\ \forall i\in[1,p]$. Therefore, the error of the constraint estimation is upper bounded \cite{xu2021crpo}, and the constraint estimation $\mathcal{J}^{i}_g(\bm{\pi})$ is precise when the utility estimation $\zeta^{i}_{K_{td}}(\bm{s},\bm{a})$ is precise. Concretely, when the utility outputs are in \textit{Real Space} (i.e., $g^i\in \mathbb{R}$), the Wasserstein distance $\overset{-}{d}_{\infty}$ (\autoref{contraction_WM}) will be degenerated to the canonical Euclidean distance whilst the distributional TD learning (\autoref{TD_learning}) will be degenerated to the traditional TD learning proposed in \cite{sutton1988learning,sutton2018reinforcement}.

\noindent \textbf{Policy Improvement}: In control setting, a distributional \textit{Bellman optimality operator} $\mathcal{T}$ with quantile approximation is proposed in \cite{dabney2018distributional}:
\begin{equation}
\mathcal{T} Z(\bm{s},\bm{a})\overset{D}{:=} R(\bm{s},\bm{a})+\gamma Z(\bm{s'},{\rm{arg}} \underset{a'}{max}\underset{\bm{p},R}{\mathbb{E}}[Z(\bm{s'},\bm{a'})])
\label{Bellman_Control_D_RL}
\end{equation}
where $\hat{Z} :=\frac{1}{N}\sum\limits_{i=1}^{N} \delta_{q_i(\bm{s},\bm{a})} \in Z_Q$ is a quantile distribution mapping one state-action pair $(\bm{s},\bm{a})$ to a uniform probability distribution supported on $q_i$. $Z_Q$ is the space of quantile distribution within $N$ supporting quantiles. $\delta_z$ denotes a Dirac with $z \in \mathbb{R}$. The state-action value $Q$ is then approximated by $Q_{j|K} \overset{D}{:=}\frac{1}{K}\sum\limits_{k=(j-1)K+1}^{(j-1)K+K}q_k(\bm{s},\bm{a})$. These quantile approximations, i.e., $\left\{q_i\right\}$, are based on Quantile Huber Loss.

We propose the ConsDRED (Algorithm~{\autoref{DRL_ConsDRED}}) to solve the CDMDP problem in \autoref{constrained_rl_problem}. The goal of ConsDRED lies in maximizing the unconstrained value function (i.e., $\mathcal{J}_r(\bm{\pi})$) whilst minimizing the constraint functions (i.e., $\mathcal{J}^{i}_g(\bm{\pi})$) if it is violated.

Let $\bm{\tau}_{c}$ be the tolerance, a newly introduced 'relaxation term' linked to constraints. Theoretically, $\bm{\tau}_{c}$  decouples the estimated cumulative constraint $\overset{-}{\mathcal{J}^{i}_g}$ to enhance the transparency of convergence. In practical terms, it serves as a lower boundary indicating the level of constraint limit that can be tolerated. The evaluation of its robustness in Section \uppercase\expandafter{\romannumeral6}-A ensures that its introduction does not cause additional tuning effort when introducing constraints.\footnote{In the benchmark \cite{xu2021crpo}, additional hyperparameters including learning rates are verified to demonstrate their robustness.} As shown in Algorithm~{\autoref{DRL_ConsDRED}} (Line 10), and we first judge whether $\overset{-}{\mathcal{J}^{i}_{g}}(\bm{\pi})\leq \bm{b}_i+\bm{\tau}_{c},\; \forall i\in[1,p]$. If so, the policy update towards minimizing $\mathcal{J}^{i}_g(\bm{\pi})$ is taken, where the approximation of the constraints is: $\overset{-}{\mathcal{J}^{i}_g}(\bm{\pi})={\mathbb{E}}[\overset{-}{Z^{i}_{t}}(\bm{s},\bm{a})]={\mathbb{E}}[\zeta^{i}_{K_{td}}(\bm{s},\bm{a})],\ \forall i\in[1,p]$; otherwise, we take the policy update towards maximizing $\mathcal{J}_r(\bm{\pi})$, where $\overset{-}{\mathcal{J}_r}(\bm{\pi})=\mathbb{E}[\overset{-}{Z^{0}_{t}}(\bm{s},\bm{a})]=\mathbb{E}[\zeta^{0}_{K_{td}}(\bm{s},\bm{a})]$. Thus, different from the traditional CMDP (i.e., \autoref{constrained_rl_problem}), the constrained distributional problem (i.e., CDMDP) is defined as:
\begin{equation}
\underset{\pi}{\rm{max}}\: \overset{-}{\mathcal{J}}_r(\bm{\pi}), \quad {\rm{s.t.}}\quad \overset{-}{\mathcal{J}^{i}_g}(\bm{\pi})\leq \bm{b}_i+\bm{\tau}_{c},\quad i=1,...,p
\label{constrained_rl_problem}
\end{equation}

\subsection{Control Parameterization}
A SADF \cite{zhang2021stochastic}, as shown in \autoref{affine_fb_SADF}, is an equivalent and tractable formulation of the original affine feedback prediction control policy proposed in \cite{jiang2001input}. Importantly, the SADF has fewer decision variables{,} which can decrease computational complexity and improve calculation efficiency.
\begin{equation}
\bm{u}_{i|t}=\sum\limits_{k=0}^{i-1} \bm{M}_{i-k|t}\bm{w}_{k|t}+\bm{v}_{i|t}
\label{affine_fb_SADF}
\end{equation}
where the $\bm{M}_{t}$ is a lower block diagonal Toeplitz structure. $i\in\mathbb{N}_{[1,N-1]}, j\in\mathbb{N}_{i-1}$ and the open-loop control sequence $\bm{v}_{i|t} \in \mathbb{R}, i\in\mathbb{N}_{N-1}$ are decision variables at each time step $t$.

According to \cite{goulart2008input}, the predicted cost can be transformed as:
\begin{equation}
\begin{aligned}
&\mathcal{L}(\bm{x}_t,\bm{u}_t)=\mathcal{L}_N(\bm{x}_t,\bm{M}_t,\bm{v}_t)\\
&=\left \|H_{x}x+H_{u}\bm{v}\right \|^{2}_2+\mathbb{E}[\left \|(H_{u}{\bm{M}}\mathcal{G}+H_{w})\bm{w}\right \|^{2}_2]
\end{aligned}
\label{predicted_cost}
\end{equation}
where $H_{x}$ and $H_{u}$ are coefficient matrices which are constructed from \autoref{predicted_cost}. $\mathcal{G}:=I_N \bigotimes G$ denotes Kronecker product of matrices $I_N$ and $G$. For the convexity guarantee, the matrix $(H_{u}{\bm{M}}\mathcal{G}+H_{w})$ is positive semidefinite (see \textit{Proposition 5} in Section \uppercase\expandafter{\romannumeral5}-B and \textit{Proposition 4.4} in \cite{goulart2007affine}).

Thus, the optimal control problem, reformulated by SADF (\autoref{affine_fb_SADF}), is as follows:
\begin{equation}
\begin{aligned}
\underset{\bm{M}_t,\bm{v}_t}{\rm{min}} \mathcal{L}_N(\bm{x}_t,\bm{M}_t&,\bm{v}_t), \quad {\rm{s.t.}}\ \forall w_{i|t}\in \mathbb{W}, \forall i\in \mathbb{N}_{N-1}\\
\rm{subject}\quad \rm{to}\quad &\bm{x}_{t} = \bm{A}x_{0|t} + \bm{B}\bm{u}_{t} + \bm{G}\bm{w}_{t}\\
&\bm{u}_{i|t}=\sum\limits_{k=0}^{i-1} \bm{M}_{i-k|t}\bm{w}_{k|t}+\bm{v}_{i|t}\\
H_{u}{\bm{M}}&\mathcal{G}+H_{w} \geq 0\qquad(\bm{x}_t,\bm{u}_t)\in \mathbb{Z}\\
\bm{x}&_{N|t}\in \mathbb{X}_{f}\qquad\qquad\bm{x}_{0|t}=\bm{x}_t\\
\end{aligned}
\label{SMPC_Formulation}
\end{equation}

The optimal control problem is a strictly convex quadratic program, or second-order cone program (SOCP) if $\mathbb{W}$ is a polytope or ellipsoid when $\mathbb{Z}$ and $\mathbb{X}_{f}$ are polytopic \cite{goulart2008input}. In this case, the problem can be seen as deterministic MPC with nonlinear constraints, which can be solved by some nonlinear MPC solvers, e.g., CasADi \cite{andersson2019casadi} and ACADOS \cite{verschueren2018towards}.

\subsection{The Whole Tracking Framework: ConsDRED-SMPC}
The objective of this work is to design a quadrotor tracking controller achieving accurate aerodynamic effect estimation, which we define as combined wind estimation and aerodynamic disturbance estimation, for tracking the trajectory reference $\bm{x}_{m,t}$ from Kino-JSS \cite{wang2022kinojgm} accurately. The immediate reward $r_{t+1}$ is defined as:
\begin{equation}
r_{t+1}=-(\bm{x}_{t}-\bm{x}_{m,t})^{\rm{T}}H_1 (\bm{x}_{t}-\bm{x}_{m,t})-\bm{u}_{t}^{\rm{T}}H_2\bm{u}_{t}
\label{reward}
\end{equation}
where $H_1$ and $H_2$ are positive definite matrices. Then, we use the DDPG architecture \cite{lillicrap2015continuous} for continuous and high-dimensional disturbance estimation.

The overall control framework for the quadrotor is shown in Fig.~\ref{RL_Control_framework}. The SADF in ConsDRED and SMPC are shown in Algorithm~{\autoref{DRL_ConsDRED}} and Algorithm~{\autoref{SADF_SMPC}}, respectively.
 
\section{Properties of ConsDRED-SMPC}
This section analyzes the properties of the proposed control framework ConsDRED-SMPC, including the convergence of ConsDRED and stability guarantees of the Controller SADF-SMPC.
\subsection{Convergence Analysis of ConsDRED}
We present \textit{Proposition 2} and \textit{Proposition 3} on the convergence analysis for the distributional RL (ConsDRED) in Section \uppercase\expandafter{\romannumeral4}-A.

\textit{Lemma 1} (\cite{bellemare2017distributional}): The \textit{Bellman operator} $\mathcal{T}^\pi$ is a $p$-contraction under the $p$-Wasserstein metric $\overset{-}{d}_{p}$.

\textit{Lemma 1} suggests that an effective practical way to minimize the Wasserstein distance between a distribution $Z$ and its Bellman update $\mathcal{T}^{\pi}Z$ can be found in \autoref{Bellman_Pred_D_RL}, which attempts iteratively to minimize the $L2$ distance between $Z$ and $\mathcal{T}^{\pi}Z$ in TD learning.

\textit{Proposition 2 (Policy Evaluation)}: Let $\Pi_{W_1}$ be a quantile approximation under the minimal 1-Wasserstein distance $W_1$, $\mathcal{T}^\pi$ be the \textit{Bellman operator} under a deterministic policy $\pi$ and $Z_{k+1}(\bm{s},\bm{a})=\Pi_{W_1}\mathcal{T}^{\pi} Z_k(\bm{s},\bm{a})$. The sequence $Z_{k}(\bm{s},\bm{a})$ converges to a unique fixed point $\overset{\sim}{Z_\pi}$ under the maximal form of $\infty$-Wasserstein metric $\overset{-}{d}_{\infty}$.

\textit{Proof}: 
\autoref{contraction_WM} implies that the combined operator $\Pi_{W_1} \mathcal{T}^\pi$ is an $\infty$-contraction \cite{dabney2018distributional}. We conclude using Banach’s fixed point theorem that $\mathcal{T}^\pi$ has a unique fixed point, i.e., $\overset{\sim}{Z_\pi}$. Furthermore, \autoref{Bellman_Control_D_RL} implies that all moments of $Z$ are bounded. Therefore, we conclude that the sequence $Z_{k}(\bm{s},\bm{a})$ converges to $\overset{\sim}{Z_\pi}$ in $\overset{-}{d}_{\infty}$ for $p\in[1,\infty]$. \hfill $\blacksquare$ 

\textit{Proposition 3 (Policy Improvement)}: Let $\bm{\pi}_{\bm{old}}$ be an old policy, $\bm{\pi}_{\bm{new}}$ be a new policy and $Q(s, a)=\mathbb{E}[Z(s,a)]$ in \autoref{Bellman_Control_D_RL}. There exists $Q^{\bm{\pi}_{\bm{new}}}(s, a) \geq Q^{\bm{\pi}_{\bm{old}}}(s, a)$, $\forall s\in \mathcal{S}$ and $\forall a\in \mathcal{A} $.

\textit{Proof}: Based on \autoref{Bellman_Control_D_RL}, there exists:
\begin{equation}
\begin{aligned}
V^\pi(s_{t})&=\mathbb{E}_\pi {Q^\pi(s_{t},\pi(s_{t}))}\leq \underset{a\in \mathcal{A}}{\rm{max}} \mathbb{E}_\pi {Q^\pi(s_{t},a)} \\
&=\mathbb{E}_{\pi'}{Q^\pi(s_{t},{\pi'}(s_{t}))}
\end{aligned}
\label{Q_V}
\end{equation}
where $\mathbb{E}_\pi[\cdot]=\sum_{a\in A}\bm{\pi}(a|s)[\cdot]$, and $V^\pi(s)=\mathbb{E}_\pi \mathbb{E}[Z_k(s,a)]$ is the value function. According to \autoref{Q_V} and \autoref{Bellman_Control_D_RL}, it yields:
\begin{equation}
\begin{aligned}
Q^{\bm{\pi}_{\bm{old}}} &=Q^{\bm{\pi}_{\bm{old}}}(s_{t},\bm{\pi}_{\bm{old}}(s_{t})) \\
&= r_{t+1}+\gamma \mathbb{E}_{s_{t+1}} \mathbb{E}_{\bm{\pi}_{\bm{old}}} Q^{\bm{\pi}_{\bm{old}}}(s_{t+1},{\bm{\pi}_{\bm{old}}}(s_{t+1}))\\
&\leq r_{t+1}+\gamma \mathbb{E}_{s_{t+1}} \mathbb{E}_{\bm{\pi}_{\bm{new}}}{Q^{\bm{\pi}_{\bm{old}}}(s_{t+1},{\bm{\pi}_{\bm{new}}}(s_{t+1}))}\\
&\leq r_{t+1}+\mathbb{E}_{s_{t+1}} \mathbb{E}_{\bm{\pi}_{\bm{new}}} [\gamma r_{t+2} \\ &+ {\gamma^2} \mathbb{E}_{s_{t+2}}{Q^{\bm{\pi}_{\bm{old}}}(s_{t+2},{\bm{\pi}_{\bm{new}}}(s_{t+2}))}|]\\
&\leq r_{t+1}+\mathbb{E}_{s_{t+1}} \mathbb{E}_{\bm{\pi}_{\bm{new}}} [\gamma r_{t+2} + {\gamma^2}r_{t+3} + ...]\\
&= r_{t+1}+\mathbb{E}_{s_{t+1}} V^{\bm{\pi}_{\bm{new}}}(s_{t+1})\\
&=Q^{\bm{\pi}_{\bm{new}}}
\end{aligned}
\label{policy_improvement}
\end{equation} \hfill $\blacksquare$

Given \textit{Proposition 2} and \textit{Proposition 3}, we can now analyze the convergence of the ConsDRED.

\textit{Theorem 1 (Global Convergence)}: Let $\bm{\pi}^{\bm{i}}$ be the policy in the $i$-th policy improvement, $i=1,2,...,\infty$, and $\bm{\pi}^{\bm{i}} \rightarrow \pi^{*}$ when $i\rightarrow\infty$. There exists $Q^{\bm  {\pi}^{*}}(s, a) \geq Q^{\bm{\pi}^{\bm{i}}}(s, a)$, $\forall s\in \mathcal{S}$ and $\forall a\in \mathcal{A}$.

\textit{Proof}: Since \textit{Proposition 3} suggests $Q^{\bm{\pi}_{\bm{i+1}}}(s, a) \geq Q^{\bm{\pi}_{\bm{i}}}(s, a)$, the sequence $Q^{\bm{\pi}_{\bm{i}}}(s, a)$ is monotonically increasing where $i \in \mathbb{N}$ is the policy iteration step. Furthermore, \textit{Lemma 1} implies that the state-action distribution $Z$ over $\mathbb{R}$ has bounded $p$-th moment, so the first moment of $Z$, i.e., $Q^{\bm{\pi}_{\bm{i}}}(s, a)$, is upper bounded. Therefore, the sequence $Q^{\bm{\pi}_{\bm{i}}}(s, a)$ converges to an upper limit $Q^{\bm{\pi}_{*}}(s, a)$ with $\forall s\in \mathcal{S}$ and $\forall a\in \mathcal{A}$. \hfill $\blacksquare$ 

Next{,} we present the convergence rate both for the \textit{TD learning} (defined in \autoref{TD_learning}) and global \textit{constrained distributional RL} (defined in \autoref{constrained_rl_problem}). Let $q_i$ be the quantile distribution supports, which equals to the outputs of the $H$-layer neural networks $f^{(h)}$: $q_i=f^{(H)}$. The $H$-layer neural networks are defined as \cite{cai2019neural}: $x^{(h)}=\frac{1}{\sqrt m} \mathbf{1}(\theta^{(h)} x^{(h-1)}>0)\cdot \theta^{(h)} x^{(h-1)},\ f^{(H)}(x=(\bm{s},\bm{a}),\theta_{K}^Q)= b_r x^{H}, \ \forall h\in[2,H]$, where the neural network parameters are bounded as $\left \|\theta \right \|_2 \leq d_{\theta}$.  The $f_{0}^{(H)}$ is defined as: $f_{0}^{(H)}=b_r x_0^{(H)},\ x_0^{(h)}((\bm{s},\bm{a}),\theta_{K}^Q)=\frac{1}{\sqrt m} \mathbf{1}(\theta^{(0)} x_{0}^{(h-1)}>0)\cdot \theta^{(h)} x_{0}^{(h-1)}$.

\textit{Proposition 4 (Convergence rate of neural TD learning)}: Let $m$ be the width of the actor-critic networks, and $\overset{-}{Z_{t}}=\frac{1}{N}\sum\limits_{i=1}^{N} \delta_{q_i(\bm{s},\bm{a})}$ be an estimator of $Z^{i}_{t}$. In TD learning, with probability at least $1-\delta$, there exists
\begin{equation}
\begin{aligned}
\left \| \Pi_{W_1}\overset{-}{Z_{t}}-\Pi_{W_1}{Z_{t}^*} \right \|& \leq \Theta(m^{-\frac{H}{4}})\\
&+\Theta({[(1-\gamma)K]^{-\frac{1}{2}}}[1+\log^{\frac{1}{2}}\delta^{-1}])
\end{aligned}
\label{convergence_rate_in_TD}
\end{equation}

\textit{Proof}: Based on Gluing lemma of Wasserstein distance $W_p$~\cite{villani2009wasserstein,clement2008elementary}, there exists:
\begin{equation}
\begin{aligned}
&\left \| \Pi_{W_1}\overset{-}{Z_{t}}-\Pi_{W_1}{Z_{t}^*} \right \|=\sum\limits_{i=1}^{N}\left \| \overset{-}{q_t^i}-q_t^{i,*} \right \| \\
&=\sum\limits_{i=1}^{N}\left \| f_i^{(H)}((\bm{s},\bm{a}),\theta_{K_{td}}^Q)-f_i^{(H)}((\bm{s},\bm{a}),\theta^{Q^*})\right \| \\
&\leq \sum\limits_{i=1}^{N}\left \| f_i^{(H)}((\bm{s},\bm{a}),\theta_{K_{td}}^Q)-f_{0,i}^{(H)}((\bm{s},\bm{a}),\theta^Q)\right \| \\
&+ \sum\limits_{i=1}^{N} \left \| f_{0,i}^{(H)}((\bm{s},\bm{a}),\theta_{K_{td}}^Q)-f_i^{(H)}((\bm{s},\bm{a}),\theta^{Q^*})\right \| \\
&\overset{(i)}{\leq}\Theta(m^{-\frac{H}{4}})+\sum\limits_{i=1}^{N} \left \| f_{0,i}^{(H)}((\bm{s},\bm{a}),\theta_{K_{td}}^Q)-f_i^{(H)}((\bm{s},\bm{a}),\theta^{Q^*})\right \| \\
&\overset{(ii)}{\leq}\Theta(m^{-\frac{H}{4}})+\Theta({[(1-\gamma)K_{td}]^{-\frac{1}{2}}}[1+\log^{\frac{1}{2}}\delta^{-1}])
\end{aligned}
\label{TD_convergence_rate_proof1}
\end{equation}
where (i) follows from \cite{cai2019neural} (\textit{Lemma 5.1}): 
\begin{equation}
\begin{aligned}
&\sum\limits_{i=1}^{N} \left \| f_i^{(H)}((\bm{s},\bm{a}),\theta_{K_{td}}^Q)-f_{0,i}^{(H)}((\bm{s},\bm{a}),\theta^Q)\right \|^2 \leq \frac{1}{m^{H}}\sum\limits_{i=1}^{N} b_r\\
&\left |[(\mathbf{1}(W_i^{(h)} x_i^{(h-1)}>0)-\mathbf{1}(W_i^{(0)} x_i^{(h-1)}>0))\cdot W_i^{(h)} x_i^{(h-1)}]^2\right | \\
&\leq \frac{4C_0}{m^{H}} \sum\limits_{i=1}^{N}[\sum\limits_{r=1}^{m}\mathbf{1}(\left | W_{i,r}^{(0)} x_i^{(h-1)}\right |\leq\left \| W_{i,r}^{(h)}-W_{i,r}^{(0)}\right \|_2 )] \\
&\leq \frac{4C_0}{m^{H}} (\sum\limits_{r=1}^{m}\left \|W_{i,r}^{(h)}-W_{i,r}^{(0)}\right \|_2^2)^{\frac{1}{2}}(\sum\limits_{r=1}^{m}\left \|\frac{1}{W_{i,r}^{(0)}}\right \|_2^2)^{\frac{1}{2}}\leq \frac{4C_0C_1}{m^{\frac{H}{2}}}
\end{aligned}
\label{TD_convergence_rate_proof2}
\end{equation}
where the constant $C_0>0$ and $C_1>0$. Thus we upper bound $\sum\limits_{i=1}^{N} \left \| f_i^{(H)}-f_{0,i}^{(H)}\right \| \leq \Theta(m^{-\frac{H}{4}})$, which holds (i) in \autoref{TD_convergence_rate_proof1}. Then (ii) follows from \cite{rahimi2008weighted} (\textit{Lemma 1}), with probability at least $1-\delta$, there exists: 
\begin{equation}
\begin{aligned}
&\sum\limits_{i=1}^{N} \left \| f_{0,i}^{(H)}((\bm{s},\bm{a}),\theta_{K_{td}}^Q)-f_i^{(H)}((\bm{s},\bm{a}),\theta^{Q^*})\right \| \\
&\leq \frac{1}{\sqrt{1-\gamma}}\sum\limits_{i=1}^{N} \left \| f_{0,i}^{(H)}((\bm{s},\bm{a}),\theta_{K_{td}}^{Q_\pi})-f_i^{(H)}((\bm{s},\bm{a}),\theta^{Q^*})\right \| \\
&\leq \frac{C_3}{\sqrt{(1-\gamma)K_{td}}}(1+\sqrt{\log{\frac{1}{\delta}}})
\end{aligned}
\label{TD_convergence_rate_proof3}
\end{equation}
where (ii) holds, and therefore \autoref{TD_convergence_rate_proof1} holds. \hfill $\blacksquare$

\textit{Proposition 4} suggests that given the deterministic hyperparameters, i.e., $H$-layer networks with width $m$ and the CDMDP discount rate $\gamma$, and after executing the TD-learning iterations in \autoref{TD_learning} for $\Theta(m^{\frac{H}{2}})$, the approximation $\overset{-}{Z_{t}}$ can be achieved by $\left \| \Pi_{W_1}\overset{-}{Z_{t}}-\Pi_{W_1}{Z_{t}^*} \right \| \leq \Theta(m^{-\frac{H}{4}})$ with high probability.

\textit{Theorem 2 (Global Convergence Rate)}:  In the training process of CDMDP (i.e., Algorithm~{\autoref{DRL_ConsDRED}}): employ the neural TD in \autoref{TD_learning} with $K_{td}=(1-\gamma)^{-\frac{3}{2}}m^{\frac{H}{2}}$, let the policy update (in \textit{Line 13} of Algorithm~{\autoref{DRL_ConsDRED}}) be $l_{Q}=\frac{1}{\sqrt{T}}$ and the tolerance be $\bm{\tau}_{c}=\Theta(\frac{1}{(1-\gamma)\sqrt{T}})+\Theta(\frac{1}{(1-\gamma)Tm^{\frac{H}{4}}})$, with probability at least $1-\delta$, there exists
\begin{equation}
\begin{aligned}
\mathcal{J}_r(\bm{\pi^*})-\mathbb{E}[\overset{-}{\mathcal{J}_r}(&\bm{\pi})]\leq\Theta(\frac{1}{(1-\gamma)\sqrt{T}})\\
&+\Theta(\frac{1}{(1-\gamma)Tm^{\frac{H}{4}}}\sqrt{\log{\frac{1}{\delta}}})
\end{aligned}
\label{Global_convergence1}
\end{equation}
Then for the constraint approximation $\overset{-}{\mathcal{J}^i_g}(\bm{\pi}),\ \forall i\in[1,p]$, there exists
\begin{equation}
\begin{aligned}
\mathbb{E}[\overset{-}{\mathcal{J}^i_g}(\bm{\pi})]-\bm{b}&_i\leq\Theta(\frac{1}{(1-\gamma)\sqrt{T}})\\
&+\Theta(\frac{1}{(1-\gamma)Tm^{\frac{H}{4}}}\sqrt{\log{\frac{1}{\delta}}})
\end{aligned}
\label{Global_convergence2}
\end{equation}

\textit{Proof}: Let $\triangle_{\theta^Q}=\theta_{t+1}^Q-\theta_{t}^Q$. Suppose the critic networks are $H$-layer neural networks. Based on \cite{kakade2002approximately} (\textit{Lemma 6.1}), there exists
\begin{equation}
\begin{aligned}
&(1-\gamma)[\mathcal{J}_r(\bm{\pi^*})-\overset{-}{\mathcal{J}_r}(\bm{\pi}_t)] \\
&=\mathbb{E}[Q_{\pi_t}(\bm{s},\bm{a})-\mathbb{E}Q_{\pi_t}(\bm{s},\bm{a}')] \\
&=\mathbb{E}[\nabla_{\theta}f^{(H)}((\bm{s},\bm{a}),\theta^Q)^{\mathrm{T}}-\mathbb{E}[\nabla_{\theta}f^{(H)}((\bm{s},\bm{a}'),\theta^Q)^{\mathrm{T}}]]\triangle_{\theta^Q}\\
&\ +\mathbb{E}[Q_{\pi_t}(\bm{s},\bm{a})-\nabla_{\theta}f^{(H)}((\bm{s},\bm{a}),\theta^Q)^{\mathrm{T}}\triangle_{\theta^Q}]\\
&\ +\mathbb{E}[\nabla_{\theta}f^{(H)}((\bm{s},\bm{a}'),\theta^Q)^{\mathrm{T}}\triangle_{\theta^Q}-Q_{\pi_t}(\bm{s},\bm{a}')]\\
&=\frac{1}{l_Q}\big[l_Q \mathbb{E}[\nabla_{\theta} \log(\bm{\pi}_t(\bm{a}|\bm{s}))^{\mathrm{T}}]\triangle_{\theta^Q}-\frac{l_{Q}^2\mathcal{L}_f}{2}\left \|\triangle_{\theta^Q}\right \|_2^2 \big]\\
&\ +\mathbb{E}[Q_{\pi_t}(\bm{s},\bm{a})-\nabla_{\theta}f^{(H)}((\bm{s},\bm{a}),\theta^Q)^{\mathrm{T}}\triangle_{\theta^Q}]+\frac{l_{Q}\mathcal{L}_f}{2}\left \|\triangle_{\theta^Q}\right \|_2^2\\
&\ +\mathbb{E}[\nabla_{\theta}f^{(H)}((\bm{s},\bm{a}'),\theta^Q)^{\mathrm{T}}\triangle_{\theta^Q}-Q_{\pi_t}(\bm{s},\bm{a}')]\\
&\overset{(i)}{\leq}\frac{1}{l_Q} \mathbb{E}[\log(\frac{\bm{\pi}_{t+1}(\bm{a}|\bm{s})}{\bm{\pi}_t(\bm{a}|\bm{s})})]+\frac{l_{Q}\mathcal{L}_f}{2}\left \|\triangle_{\theta^Q}\right \|_2^2\\
&\ +\sqrt{\mathbb{E}[Q_{\pi_t}(\bm{s},\bm{a})-f^{(H)}((\bm{s},\bm{a}),\triangle_{\theta^Q})]^2}\\
&\ +\sqrt{\mathbb{E}[f^{(H)}((\bm{s},\bm{a}),\triangle_{\theta^Q})-\nabla_{\theta}f^{(H)}((\bm{s},\bm{a}),\theta^Q)^{\mathrm{T}}\triangle_{\theta^Q}]^2}\\
&\ +\sqrt{\mathbb{E}[\nabla_{\theta}f^{(H)}((\bm{s},\bm{a}'),\theta^Q)^{\mathrm{T}}\triangle_{\theta^Q}-f^{(H)}((\bm{s},\bm{a}'),\triangle_{\theta^Q})]^2} \\
&\ +\sqrt{\mathbb{E}[f^{(H)}((\bm{s},\bm{a}'),\triangle_{\theta^Q})-Q_{\pi_t}(\bm{s},\bm{a}')]^2}\\
&\ =\frac{1}{l_Q}\big[\mathbb{E}[\mathcal{D}_{KL}(\bm{\pi^*}||\bm{\pi}_{t})]-\mathbb{E}[\mathcal{D}_{KL}(\bm{\pi^*}||\bm{\pi}_{t+1})]\big]\\
&\ +2\sqrt{\mathbb{E}[f^{(H)}((\bm{s},\bm{a}),\triangle_{\theta^Q})-\nabla_{\theta}f^{(H)}((\bm{s},\bm{a}),\theta^Q)^{\mathrm{T}}\triangle_{\theta^Q}]^2}\\
&\ +2\sqrt{\mathbb{E}[Q_{\pi_t}(\bm{s},\bm{a})-f^{(H)}((\bm{s},\bm{a}),\triangle_{\theta^Q})]^2}+\frac{l_{Q}\mathcal{L}_f}{2}\left \|\triangle_{\theta^Q}\right \|_2^2
\end{aligned}
\label{Global_convergence_proof1}
\end{equation}
where (i) follows from the $\mathcal{L}_f$-Lipschitz property of $\log(\bm{\pi}_t(\bm{a}|\bm{s}))$. Next, we upper bound the term $\sqrt{\mathbb{E}[f^{(H)}((\bm{s},\bm{a}),\triangle_{\theta^Q})-\nabla_{\theta}f^{(H)}((\bm{s},\bm{a}),\theta^Q)^{\mathrm{T}}\triangle_{\theta^Q}]^2}$ as shown below.
\begin{equation}
\begin{aligned}
&\sqrt{\mathbb{E}[f^{(H)}((\bm{s},\bm{a}),\triangle_{\theta^Q})-\nabla_{\theta}f^{(H)}((\bm{s},\bm{a}),\theta^Q)^{\mathrm{T}}\triangle_{\theta^Q}]^2}\\
&=\sum\limits_{i=1}^{N}\left \| f_i^{(H)}((\bm{s},\bm{a}),\triangle_{\theta^Q})-\nabla_{\theta}f_i^{(H)}((\bm{s},\bm{a}),\theta^Q)^{\mathrm{T}}\triangle_{\theta^Q}\right \| \\
&\leq \sum\limits_{i=1}^{N}\big[ \left \| f_i^{(H)}((\bm{s},\bm{a}),\triangle_{\theta^Q})-\nabla_{\theta}f_{0,i}^{(H)}((\bm{s},\bm{a}),\theta^Q)^{\mathrm{T}}\triangle_{\theta^Q} \right \| \\
&+\left \|\nabla_{\theta}f_{0,i}^{(H)}((\bm{s},\bm{a}),\theta^Q)^{\mathrm{T}}\triangle_{\theta^Q}-\nabla_{\theta}f_i^{(H)}((\bm{s},\bm{a}),\theta^Q)^{\mathrm{T}}\triangle_{\theta^Q}\right \| \big]\\
&=2\sum\limits_{i=1}^{N}\left \| f_i^{(H)}((\bm{s},\bm{a}),\triangle_{\theta^Q})-f_{0,i}^{(H)}((\bm{s},\bm{a}),\triangle_{\theta^Q})\right \| \\
&\overset{(ii)}{\leq} \frac{4\sqrt{C_0C_1}}{m^{\frac{H}{4}}}
\end{aligned}
\label{Global_convergence_proof2}
\end{equation}
where (ii) follows from \autoref{TD_convergence_rate_proof2}. Then, in order to upper bound $\sqrt{\mathbb{E}[Q_{\pi_t}(\bm{s},\bm{a})-f^{(H)}((\bm{s},\bm{a}),\triangle_{\theta^Q})]^2}$, taking expectation of \autoref{Global_convergence_proof1} from $t=0$ to $T-1$, yields
\begin{equation}
\begin{aligned}
&(1-\gamma)\big[\mathcal{J}_r(\bm{\pi^*})-\mathbb{E}[\overset{-}{\mathcal{J}_r}(\bm{\pi})]\big]\\
&=(1-\gamma)\frac{1}{T}\sum\limits_{t=0}^{T-1}[\mathcal{J}_r(\bm{\pi^*})-\overset{-}{\mathcal{J}_r}(\bm{\pi}_t)]\\
&\leq \frac{1}{T}\big[\frac{1}{l_Q}\mathbb{E}[\mathcal{D}_{KL}(\bm{\pi^*}||\bm{\pi}_{t})]+\frac{8T\sqrt{C_0C_1}}{m^{\frac{H}{4}}}+\frac{Tl_{Q}\mathcal{L}_f}{2}d_{\theta}^2\\
&+2\sum\limits_{t=0}^{T-1}\sum\limits_{i=1}^{N}\left \| f_i^{(H)}((\bm{s},\bm{a}),\theta_{t+1}^Q-\theta_{t}^Q)-f_i^{(H)}((\bm{s},\bm{a}),\theta^{Q^*})\right \|\big] \\
&= \frac{\mathbb{E}[\mathcal{D}_{KL}(\bm{\pi^*}||\bm{\pi}_{t})]}{l_QT}+\frac{8\sqrt{C_0C_1}}{m^{\frac{H}{4}}}+\frac{l_{Q}\mathcal{L}_f}{2}d_{\theta}^2\\
&\ +\frac{2}{T}\sum\limits_{i=1}^{N}\left \| f_i^{(H)}((\bm{s},\bm{a}),\theta_{K_{td},t}^Q)-f_i^{(H)}((\bm{s},\bm{a}),\theta^{Q^*})\right \| \\
&\overset{(iii)}{\leq} \frac{\mathbb{E}[\mathcal{D}_{KL}(\bm{\pi^*}||\bm{\pi}_{t})]}{l_QT}+\frac{8\sqrt{C_0C_1}}{m^{\frac{H}{4}}}+\frac{l_{Q}\mathcal{L}_f}{2}d_{\theta}^2\\
&\ +\frac{4\sqrt{C_0C_1}}{Tm^{\frac{H}{4}}}+\frac{2C_3}{T\sqrt{(1-\gamma)K_{td}}}(1+\sqrt{\log{\frac{1}{\delta}}})
\end{aligned}
\label{Global_convergence_proof3}
\end{equation}
where (iii) follows from \textit{Proposition 4} (\autoref{TD_convergence_rate_proof1}). Thus, substituting $K_{td}=(1-\gamma)^{-1}m^{\frac{H}{2}}$ and $l_{Q}=\Theta(1/\sqrt{T})$ into \autoref{Global_convergence_proof3}, with probability at least $1-\delta$, yields: 
\begin{equation}
\begin{aligned}
&\mathcal{J}_r(\bm{\pi^*})-\mathbb{E}[\overset{-}{\mathcal{J}_r}(\bm{\pi})]\leq C_5\frac{1}{(1-\gamma)\sqrt{T}}+C_6\frac{1}{(1-\gamma)m^{\frac{H}{4}}}\\
&\ +C_7\frac{1}{(1-\gamma)Tm^{\frac{H}{4}}}+2C_3\frac{\sqrt{\log{\frac{1}{\delta}}}}{(1-\gamma)Tm^{\frac{H}{4}}}\\
&\leq\Theta(\frac{1}{(1-\gamma)\sqrt{T}})+\Theta(\frac{1}{(1-\gamma)Tm^{\frac{H}{4}}}\sqrt{\log{\frac{1}{\delta}}})
\end{aligned}
\label{Global_convergence_proof4}
\end{equation}
where $C_5=\mathbb{E}[\mathcal{D}_{KL}(\bm{\pi^*}||\bm{\pi}_{t})]+\frac{\mathcal{L}_fd_{\theta}^2}{2}$, $C_6=8\sqrt{C_0C_1}$ and $C_7=4\sqrt{C_0C_1}+2C_3$. Therefore, \autoref{Global_convergence1} holds.

Following \textit{Line 13} in Algorithm~\autoref{DRL_ConsDRED} and recalling \autoref{Global_convergence_proof1}, \autoref{Global_convergence_proof2} and \autoref{Global_convergence_proof3}, the convergence process is similarly stated for the constraint approximation $\overset{-}{\mathcal{J}^{i}_g}(\bm{\pi}),\ \forall i\in[1,p]$ here
\begin{equation}
\begin{aligned}
\mathbb{E}[\overset{-}{\mathcal{J}^i_g}(\bm{\pi})]-\mathcal{J}^i_g(\bm{\pi^*})&\leq\Theta(\frac{1}{(1-\gamma)\sqrt{T}})\\
&\ +\Theta(\frac{1}{(1-\gamma)Tm^{\frac{H}{4}}}\sqrt{\log{\frac{1}{\delta}}})
\end{aligned}
\label{Global_convergence_proof5}
\end{equation}
the constraint violation is then bounded below
\begin{equation}
\begin{aligned}
&\mathbb{E}[\overset{-}{\mathcal{J}^i_g}(\bm{\pi})]-\bm{b}_i \leq \big[\mathcal{J}^i_g(\bm{\pi^*})-\bm{b}_i\big]+\big[\mathbb{E}[\overset{-}{\mathcal{J}^i_g}(\bm{\pi})]-\mathcal{J}^i_g(\bm{\pi^*})\big]\\
&\leq \bm{\tau}_{c}+\big[\mathbb{E}[\overset{-}{\mathcal{J}^i_g}(\bm{\pi})]-\mathcal{J}^i_g(\bm{\pi^*})\big]\\
&\leq \bm{\tau}_{c}+\Theta(\frac{1}{(1-\gamma)\sqrt{T}})+\Theta(\frac{1}{(1-\gamma)Tm^{\frac{H}{4}}}\sqrt{\log{\frac{1}{\delta}}})
\end{aligned}
\label{Global_convergence_proof6}
\end{equation}
where we have $\bm{\tau}_{c}=\Theta(\frac{1}{(1-\gamma)\sqrt{T}})+\Theta(\frac{1}{(1-\gamma)Tm^{\frac{H}{4}}})$, therefore, \autoref{Global_convergence2} holds: $\mathbb{E}[\overset{-}{\mathcal{J}^i_g}(\bm{\pi})]-\bm{b}_i\leq\Theta(\frac{1}{(1-\gamma)\sqrt{T}})+\Theta(\frac{1}{(1-\gamma)Tm^{\frac{H}{4}}}\sqrt{\log{\frac{1}{\delta}}})$. \hfill $\blacksquare$ 

According to \textit{Theorem 2}, \textbf{1)} CDMDP (i.e., Algorithm~\autoref{DRL_ConsDRED}) guarantees convergence to the global optimal policy $\bm{\pi^*}$ at a sublinear rate $\Theta(1/{\sqrt{T}})$ whilst achieving an approximation error $\Theta(1/{m^{\frac{H}{4}}})$ decrease as the width and the layer of neural network $m$ and $H$ increase; and \textbf{2)} the constraint violation (shown in \autoref{Global_convergence2}) also converges to \textit{zero} at a sublinear rate $\Theta(1/{\sqrt{T}})$ with an error $\Theta(1/{m^{\frac{H}{4}}})$ decrease as $m$ and $H$ increase. Therefore, to obtain an output policy $\bm{\pi}_{out}$ reaching $\mathcal{J}_r(\bm{\pi^*})-\mathbb{E}[\overset{-}{\mathcal{J}_r}(\bm{\pi}_{out})]\leq\xi$ and $\mathbb{E}[\overset{-}{\mathcal{J}^i_g}(\bm{\pi}_{out})]-\bm{b}_i\leq \xi$, CDMDP needs at most $T=\Theta(\xi^{-2})$ iterations.

\subsection{Stability Guarantee of the Controller}
This subsection will demonstrate the closed-loop stability of {the} ConsDRED-SMPC control framework. The closed-loop stability is analyzed under the Lipschitz Lyapunov function \cite{rifford2000existence} to guarantee ISS. Before the closed-loop stability analysis, the convexity and Lipschitz continuity of the cost function $\mathcal{L}_{\bm{N}}(\bm{x}_t,\bm{M}_t,\bm{v}_t)$ are introduced in \textit{Proposition 5} and \textit{Proposition 6}, respectively. Since the output of ConsDRED is non-zero-mean and bounded values, which are different from the assumption of zero-mean disturbances in most previous work (\cite{zhang2021stochastic} and \cite{goulart2008input}), the following proofs are all based on the non-zero-mean and bounded disturbances.

We first define an optimal control policy based on the affine disturbance feedback control law:
\begin{equation}
(\bm{M}^{*}(x),\bm{v}^{*}(x)):=\arg \underset{(\bm{M},\bm{v})\in \mathcal{V}_N}{\rm{min}} \mathcal{L}_N(\bm{x},\bm{M},\bm{v})
\label{optimal_control_policy}
\end{equation}
where $\mathcal{V}_N$ is the set of feasible policies, and $(\bm{M}^{*}(x),\bm{v}^{*}(x))$ is a optimal control policy group. The optimal value function $\mathcal{L}^{*}_N(x)$ under the affine disturbance feedback control law is defined as:
\begin{equation}
\mathcal{L}^{*}_N(x):=\underset{(\bm{M},\bm{v})\in \mathcal{V}_N}{\rm{min}} \mathcal{L}_N(\bm{x},\bm{M},\bm{v})
\label{optimal_value}
\end{equation}

Then we demonstrate that the optimal value function $\mathcal{L}_N(x)$ is convex (see \textit{Proposition 5}) so that  \autoref{optimal_value} can be operated as a convex optimization problem.

\textit{Proposition 5}: The function $\mathcal{L}_N(\bm{x},\bm{M},\bm{v})$ is convex.

\textit{Proof}: In \autoref{predicted_cost}, the second term  $\mathbb{E}[\left \|(H_{u}{\bm{M}}\mathcal{G}+H_{w})\bm{w}\right \|^{2}_2]$, i.e., the expected value of a quadratic form with respect to the vector-valued random variable $\bm{w}$, is equal to:
\begin{equation}
\begin{aligned}
&\mathbb{E}[\left \|(H_{u}{\bm{M}}\mathcal{G}+H_{w})\bm{w}\right \|^{2}_2]=\mathbb{E}[{\rm{tr}}((H_{u}{\bm{M}}\mathcal{G}+H_{w})\bm{w} \bm{w}^T)]\\
&={\rm{tr}}((H_{u}{\bm{M}}\mathcal{G}+H_{w})\mathbb{E}[\bm{w} \bm{w}^T])\\
&={\rm{tr}}((H_{u}{\bm{M}}\mathcal{G}+H_{w})({\rm{Cov}}(\bm{w})+\bm{\mu} \bm{\mu}^T))\\
&={\rm{tr}}(\bm{C}^{\frac{1}{2}}_{\bm{w}}(H_{u}{\bm{M}}\mathcal{G}+H_{w})^{T}(H_{u}{\bm{M}}\mathcal{G}+H_{w})\bm{C}^{\frac{1}{2}}_{\bm{w}})\\
&+\bm{\mu}^T(H_{u}{\bm{M}}\mathcal{G}+H_{w})\bm{\mu}
\end{aligned}
\label{convex_simplied_w}
\end{equation}
where $\rm{tr}(\cdot)$ denotes the trace of a square matrix. $\bm{\mu}=\mathbb{E}(\bm{w})$ is the expected value of $\bm{w}$, and $\bm{C}_{\bm{w}}={\rm{Var}}(\bm{w})$ is the variance-covariance matrix of $\bm{w}$. Therefore, $\mathcal{L}_N(x)$ can be written as:
\begin{equation}
\begin{aligned}
&\mathcal{L}_N(x)=\left \|H_{x}x+H_{u}\bm{v}\right \|^{2}_2+\left \|\bm{\mu}\right \|^{2}_{(H_{u}{\bm{M}}\mathcal{G}+H_{w})}\\
&+{\rm{tr}}(\bm{C}^{\frac{1}{2}}_{\bm{w}}(H_{u}{\bm{M}}\mathcal{G}+H_{w})^{T}(H_{u}{\bm{M}}\mathcal{G}+H_{w})\bm{C}^{\frac{1}{2}}_{\bm{w}})
\end{aligned}
\label{convex_l_n}
\end{equation}
where $\left \|x\right \|_{P}$ denotes weighted $2$–norm of the vector $x$. \autoref{convex_l_n} is convex since it consists of convex functions of vector and matrix norms.\hfill $\blacksquare$ 

\textit{Proposition 6}: The function $\mathcal{L}^{*}_N(\bm{x},\bm{M},\bm{v})$ is Lipschitz continuous.

\textit{Proof}: The cost function $\mathcal{L}_N(\bm{x},\bm{M},\bm{v})$ is proved to be convex in \textit{Proposition 5} so that $\mathcal{L}^{*}_N(\bm{x},\bm{M},\bm{v})$ is convex if $\mathcal{V}_N$ has a non-empty interior (\textit{Proposition 1} of \cite{goulart2008input}). $Z$ is a compact (closed and bounded) set so that $\mathcal{L}^{*}_N(\bm{x},\bm{M},\bm{v})$, defined under the compact space $Z$, is piecewise quadratic (\textit{Corollary 4.6} of \cite{goulart2007affine}). Therefore $\mathcal{L}^{*}_N(\bm{x},\bm{M},\bm{v})$ is a Lipschitz continuity function. \hfill $\blacksquare$

The above results lead directly to our final result:

\textit{Theorem 3}: Let $\mathcal{W}$, $\mathcal{Z}$ and $\mathcal{X}_{f}$ be polytopes. The closed-loop system (\autoref{linear_discrete_system}) under the SADF control law $\bm{u}_{i|t}$ (in \autoref{affine_fb_SADF}) is ISS. The ISS is also guaranteed in such cases: the stochastic disturbance $\bm{w}_t$ is an i.i.d. bounded and non-zero-mean distribution, i.e., $\mathbb{E}{(w_k)}\neq0$.

\textit{Proof}: According to \textit{Proposition 5} and \textit{Proposition 6}, we first state that the optimal value function $\mathcal{L}^{*}_N(x)$ is a Lipschitz continuity function. The key is then to prove that there exists a Lipschitz continuous function, i.e., $\mathcal{L}^{*}_N(x)$, to satisfy the Lipschitz-ISS criterion (\textit{Proposition 4.15} in \cite{goulart2008input}).

According to \textit{Proposition 1}, there exists a baseline control law $\bm{u}_b$ ensuring ISS under zero-mean distribution disturbance. Let $V_b (x)=V^*_{Nb}(x)-V^*_{Nb}(0)$ be the Lipschitz continuous Lyapunov function \cite{goulart2008input}, where $V^*_{Nb}(x)$ is the optimal value function under the baseline control law $\bm{u}_b$. Here let $x^{+}=f(x,w)$ \cite{goulart2008input,munoz2020convergence}, so there exists:
\begin{subequations}
\begin{align}
\alpha_1(\left \| x \right\|) \leq &V_b(x) \leq \alpha_2(\left \| x \right\|) \label{ISS_baseline_control_law_a}\\
V_b(f(x,0)) - &V_b(x) \leq -\alpha_3(\left \| x \right\|) \label{ISS_baseline_control_law_b}
\end{align}
\end{subequations}

\begin{algorithm}[t]
\caption{SADF-SMPC}
\begin{algorithmic}[1]
\STATE \textbf{Get}:\\
    - the reference data $\bm{x}_{ref}$ from the quadrotor trajectory planning and generation module, i.e., Kino-JSS \cite{wang2022kinojgm}\\
    - the measurement state $\bm{x}_k$ from on-board depth camera and IMU (see \autoref{hardware_spec} for details please)\\
    - the wind estimation $\bm{e}_{fk}$ from VID-Fusion \cite{ding2020vid}\\
\STATE \textbf{Initialize}:\\
    - the parameters $\theta^\mu$ and $\theta^Q$ for the actor $\mu$ and the critic $Q$, respectively\\
    - the decision variables $\bm{M}_{0}$ and $\bm{v}_{0}$ in \autoref{affine_fb_SADF}\\
    - the initial state $\bm{s}_{0}$\\
\FOR{each sampling timestamps $k$}
\STATE \textbf{Repeat}
\STATE $\bm{s}_k \gets \left [ \bm{x}_k, \bm{e}_{fk} \right ]$
\STATE Select an action vector $\bm{w}_k \gets [w_{0|k}^{\rm{T}}, w_{1|k}^{\rm{T}}, ..., w_{N|k}^{\rm{T}}]^{\rm{T}}$ from $\bm{w}_k=\bm{a}_k \gets \mu(\bm{a}_k|\bm{s}_k)$ in ConsDRED (Algorithm~\autoref{DRL_ConsDRED})
\STATE $\bm{u}_{i|k} \gets \sum\limits_{l=0}^{i-1} \bm{M}_{i-l|k}\bm{w}_{l|k}+\bm{v}_{i|k}$
\STATE $\bm{u}_{k} \gets \bm{u}_{0|k}$, $\bm{w}_{k} \gets \bm{w}_{0|k}$
\STATE $\bm{x}_{k} \gets \bm{A}x_{0|k} + \bm{B}\bm{u}_{k} + \bm{G}\bm{w}_{k}$
\STATE Solve optimization problem \autoref{SMPC_Formulation} with nonlinear MPC solver, i.e., ACADOS \cite{verschueren2018towards}\\
\STATE \textbf{Until} convergence
\STATE $\bm{u}_{k} \gets \bm{v}_{0|k}$
\STATE $\bm{x}_{k+1}$, $\bm{e}_{fk+1} \gets {\rm{Real Quadrotor}}(\bm{u}_{k})$
\STATE $\bm{s}_{k+1} \gets [\bm{x}_{k+1}, \bm{e}_{fk+1}]$
\STATE $\bm{x}_{ref} \gets$ Kino-JSS
\STATE $k \gets k+1$
\ENDFOR
\end{algorithmic}
\label{SADF_SMPC}
\end{algorithm}

\begin{algorithm}[t]
\caption{ConsDRED}
\hspace*{0.02in} {\bf Input:}
$\bm{s}_k$, $\bm{s}_{k+1}$, $\bm{u}_{k}$, $\theta^\mu$, $\theta^Q$\\
\hspace*{0.02in} {\bf Output:}
$\bm{a}_k$
\begin{algorithmic}[1]
\STATE \textbf{Initialize}:\\
    - $\bm{\theta}=[\theta^\mu,\theta^Q]$: the parameters of {the} actor network $\theta^\mu$, and the parameters of {the} critic network $\theta^Q$\\
    - the replay memory $D \gets D_{k-1}$\\
    - the batch $B$, and its size\\
    - the random option selection probability $\epsilon$\\
    - the option termination probability $\beta$ \\
\STATE \textbf{Repeat}
\FOR{$t=0,1,2,...,T-1$}
\STATE Select action $\bm{a}_t=\bm{\pi}(\bm{s}_t|\bm{\theta^\mu}+{\mathcal{N}_t})$ based on the current policy and exploration Gaussian noise
\STATE Execute action $\bm{a_t}$ (i.e., $\bm{w}_k$ in Algorithm~{\autoref{SADF_SMPC}}), get reward $r_t$ and the next state $\bm{s}_{t+1}$
\STATE $D.{\bf{insert}}([{\bm{s}}_{k}, a_{k}, r_k, {\bm{s}}_{k+1}])$, and $([{\bm{s}_n}, a_n, r_n, {\bm{s}}_{n+1}])$ of $N$ tuples $ \in B \gets D.\bf{sampling}$
\STATE \textbf{policy evaluation}: $z_{t,n}^{i} \gets \zeta^{i}_{K_{td}}(\bm{s},\bm{a})$, $\forall i\in[0,p]$
\STATE \textbf{constraint estimation}: compute the constraint estimation: $\overset{-}{\mathcal{J}^{i}_g}(\bm{\pi_{\theta}}(\bm{s},\bm{a}))={\mathbb{E}}[\zeta^{i}_{K_{td}}(\bm{s},\bm{a})]$, $\ \forall i\in[1,p]$\\
\STATE \textbf{policy improvement}:\\
\IF{$\overset{-}{\mathcal{J}^{i}_{g}}(\bm{\pi})\leq \bm{b}_i+\bm{\tau}_{c}$}
\STATE Update the policy towards maximizing $\mathcal{J}_r(\bm{\pi})$:\\
compute the actor update: $\delta_{\theta^\mu} \gets (1/N)\sum\limits_{n=0}{\nabla_{\theta^\mu}}{\pi_{\theta^\mu}}(\bm{s}_n)\mathbb{E}[{\nabla _a}Z_{\theta^Q}(\bm{s}_n,\bm{a})]_{a=\pi_{\theta^\mu}(\bm{s}_n)}$ \\
compute the critic update: $\delta_{\theta^Q} \gets (1/N)\sum\limits_{n=0} \nabla_{\theta^Q} \overset{-}{d}_{\infty}(\Pi_{W_1}\mathcal{T}^{\pi}Z_{\theta^Q}(\bm{s}_n,\bm{a}_n),\Pi_{W_1}\mathcal{T}^{\pi}z_{t,n}^{0})$\\
$\theta^\mu \gets \theta^\mu + l_{\mu}\delta_{\theta^\mu}$, and $\theta^Q \gets \theta^Q + l_{\theta} \delta_{\theta^Q}$
\ELSE
\STATE Update the policy towards minimizing $\mathcal{J}^{i}_g(\bm{\pi})$:\\
$\theta^\mu \gets \theta^\mu - l_{\mu}\nabla_{\theta^\mu} \zeta^{i}_{K_{td}}$, and $\theta^Q \gets \theta^Q - l_{Q} \nabla_{\theta^Q} \zeta^{i}_{K_{td}}$, $\forall i\in[1,p]$
\ENDIF
\ENDFOR
\end{algorithmic}
\label{DRL_ConsDRED}
\end{algorithm}

Let $V(x)=\mathcal{L}^*_{N}(x)-\mathcal{L}^*_{N}(0)$, where $\mathcal{L}^*_{N}(x)$ is an optimal value function under the affine disturbance feedback control law with bounded and non-zero-mean disturbance distribution (\autoref{optimal_value}). According to \autoref{convex_l_n}, $L^*_{N}(x)$ is shown as:
\begin{equation}
\begin{aligned}
&\mathcal{L}^*_{N}(x)={\rm{min}} \left.\{\mathcal{L}_{N}(x) \right.\}\\
&={\rm{min}}\left.\{ \left \|H_{x}x+H_{u}\bm{v}\right \|^{2}_2 +\left \|\bm{\mu}\right \|^{2}_{(H_{u}{\bm{M}}\mathcal{G}+H_{w})} \right.\\
&\left.+{\rm{tr}}(\bm{C}^{\frac{1}{2}}_{\bm{w}}(H_{u}{\bm{M}}\mathcal{G}+H_{w})^{T}(H_{u}{\bm{M}}\mathcal{G}+H_{w})\bm{C}^{\frac{1}{2}}_{\bm{w}}) \right.\}\\
&=V^*_{Nb}(x)+{\rm{min}}\left.\{\left \|\bm{\mu}\right \|^{2}_{(H_{u}{\bm{M}}\mathcal{G}+H_{w})} \right.\}
\end{aligned}
\label{ISS_opyimal_value}
\end{equation}
where $\bm{\mu}$ is the expected value of disturbances, which is independent with $\bm{v}$ and $\bm{M}$. $V(x)=\mathcal{L}^*_{N}(x)-\mathcal{L}^*_{N}(0)=V^*_{Nb}(x)-V^*_{Nb}(0)=V_b (x)$. Hence, there exists $\mathcal{H}_{\infty}$-functions $\alpha_1(\cdot)$, $\alpha_2(\cdot)$ such that \autoref{ISS_baseline_control_law_a} holds with $V_b(\cdot)=V(\cdot)=\mathcal{L}^*_{N}(\cdot)-\mathcal{L}^*_{N}(0)$.

To prove $V(\cdot)$ satisfying \autoref{ISS_baseline_control_law_b}, note that $V_b(f(x,0)) - V_b(x)=[V^*_{Nb}(f(x,0))-V^*_{Nb}(0)]-[V^*_{Nb}(x)-V^*_{Nb}(0)]=V^*_{Nb}(f(x,0))-V^*_{Nb}(x)$, so that $V^*_{Nb}(f(x,0))-V^*_{Nb}(x) \leq -\alpha_3(\left \| x \right\|)$. It follows that:
\begin{equation}
\begin{aligned}
V&(f(x,0)) - V(x) \\
&= [\mathcal{L}^*_{N}(f(x,0))-\mathcal{L}^*_{N}(0)]-[\mathcal{L}^*_{N}(x)-\mathcal{L}^*_{N}(0)]\\
&=\mathcal{L}^*_{N}(f(x,0))-\mathcal{L}^*_{N}(x)
\end{aligned}
\label{ISS_V_second_inequality}
\end{equation}
where both $\mathcal{L}^*_{N}(f(x,0))$ and $V^*_{Nb}(f(x,0))$ have $w=0$. The only difference between the two control laws is zero-mean or non-zero-mean disturbance distributions so that $\mathcal{L}^*_{N}(f(x,0))=V^*_{Nb}(f(x,0))$. Hence, combining with \autoref{ISS_opyimal_value}, \autoref{ISS_V_second_inequality} can be rewritten as:
\begin{equation}
\begin{aligned}
V&(f(x,0)) - V(x) \\
&=V^*_{Nb}(f(x,0))-V^*_{Nb}(x)-{\rm{min}}\left.\{ \left \|\bm{\mu}\right \|^{2}_{(H_{u}{\bm{M}}\mathcal{G}+H_{w})} \right.\}
\end{aligned}
\label{ISS_V_second_inequality1}
\end{equation}

According to \textit{Proposition 5}, we have ${\rm{min}}\left.\{\left \|\bm{\mu}\right \|^{2}_{(H_{u}{\bm{M}}\mathcal{G}+H_{w})}\right.\} \geq 0$. Then we have:
\begin{equation}
\begin{aligned}
V&(f(x,0)) - V(x) \\
&\leq V^*_{Nb}(f(x,0))-V^*_{Nb}(x) \leq -\alpha_3(\left \| x \right\|)
\end{aligned}
\label{ISS_V_second_inequality2}
\end{equation}
\autoref{ISS_V_second_inequality2} above shows that there exists $\mathcal{H}_{\infty}$-functions $\alpha_3(\cdot)$ such that \autoref{ISS_baseline_control_law_b} holds with $V(\cdot)=\mathcal{L}^*_{N}(\cdot)-\mathcal{L}^*_{N}(0)$. Therefore, $V(x)=\mathcal{L}^*_{N}(x)-\mathcal{L}^*_{N}(0)$ is a Lipschitz continuous Lyapunov function, and the ISS of the closed-loop system (\autoref{linear_discrete_system}) is guaranteed with bounded and non-zero-mean distribution disturbances, i.e., $\mathbb{E}{(w_k)}\neq0$. \hfill $\blacksquare$ 

\section{Numerical Example}
This section shows the whole training process in both RotorS \cite{furrer2016rotors}, a UAV gazebo simulator \footnote{We simulate external winds in Gazebo by incorporating a plugin file. The rationale behind this simulation is to emulate the forces exerted on a body, assuming them as external aerodynamic effects.}, and real-world scenarios, where empirical convergence results are consistent with our understanding from theoretical analysis in Section \uppercase\expandafter{\romannumeral5}-A. Then, the comparative performance of our proposed ConsDRED-SMPC is evaluated in both simulated and real-world flight experiments, respectively.

\begin{table}[t]
    \caption{Parameters of ConsDRED-SMPC}
    \label{model_parameters}
    \setlength{\tabcolsep}{1.5mm}{
    \begin{tabular}{c c c}
    \toprule
    \textbf{Parameters} & \textbf{Definition} & \textbf{Values} \\
    \hline
    $l_{\mu}$ & Learning rate of actor & 0.001 \\
    $l_{\theta}$ & Learning rate of critic & 0.001 \\
    $\mu$ & \makecell[c]{Actor neural network: fully connected with $H$ \\hidden layers ($m$ neurons per hidden layer)} & - \\
    $\theta$ & \makecell[c]{Critic neural network: fully connected with $H$ \\hidden layers ($m$ neurons per hidden layer)} & - \\
    $D$ & Replay memory capacity & $10^6$ \\
    $B$ & Batch size & 128 \\
    $\gamma$ & Discount rate & 0.998 \\
    - & Training episodes & 1000 \\ 
    $m$ & the width of neural network & 32 \\
    $H$ & the layer of neural network & 2 \\
    $T$ & Length in each episode & 200 \\
    $T_s$ & MPC Sampling period & 50ms \\
    $N$ & Time steps & 20 \\
    \bottomrule
    \end{tabular}}
\end{table}

\begin{figure}[t]
  \centering
  \includegraphics[scale=0.4]{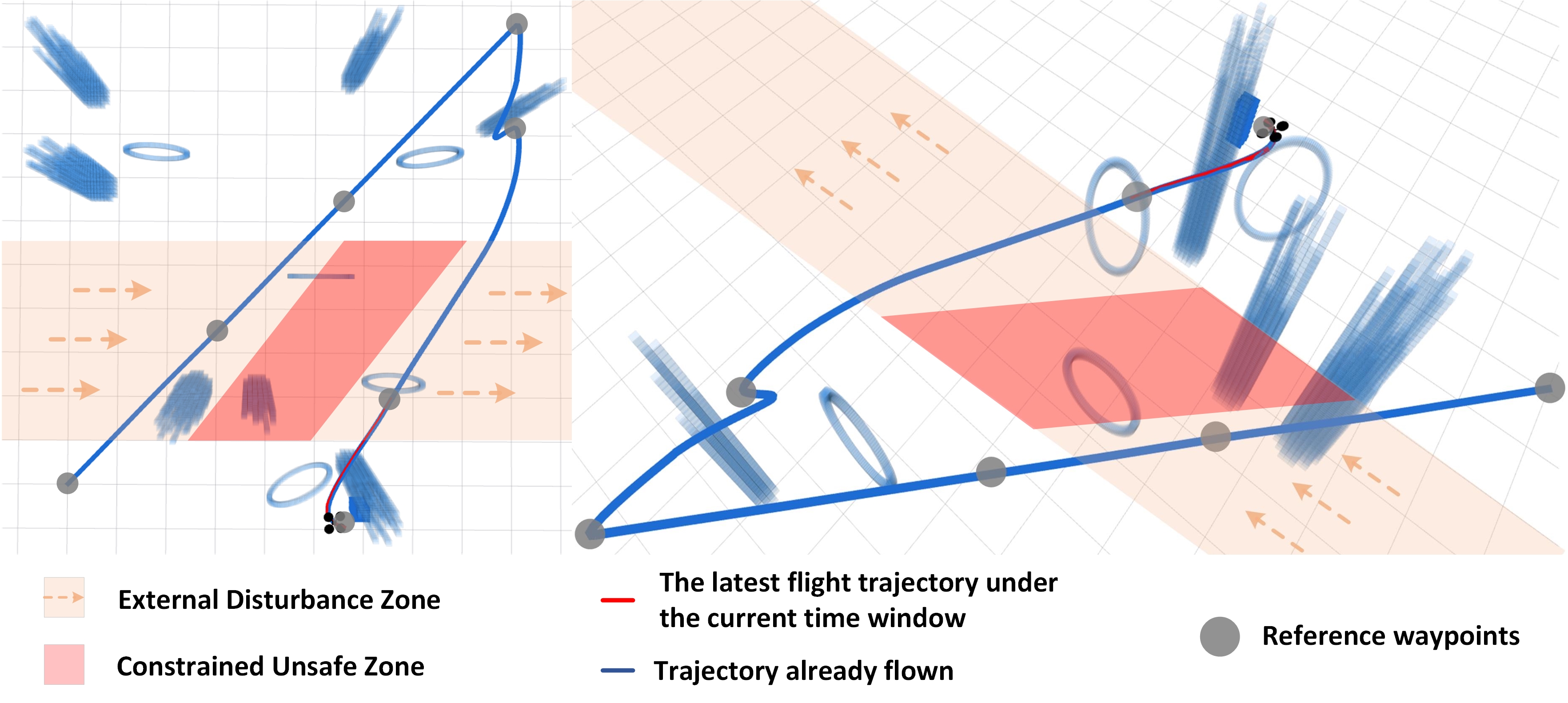}
  \caption{The simulation scenario: both reference trajectories with/without external forces are generated by Kino-JSS \cite{wang2022kinojgm}.}
  \label{simulation_env}
\end{figure}

\subsection{Comparative performance of ConsDRED Training}
\begin{figure}[t]
  \centering
  \includegraphics[scale=0.19]{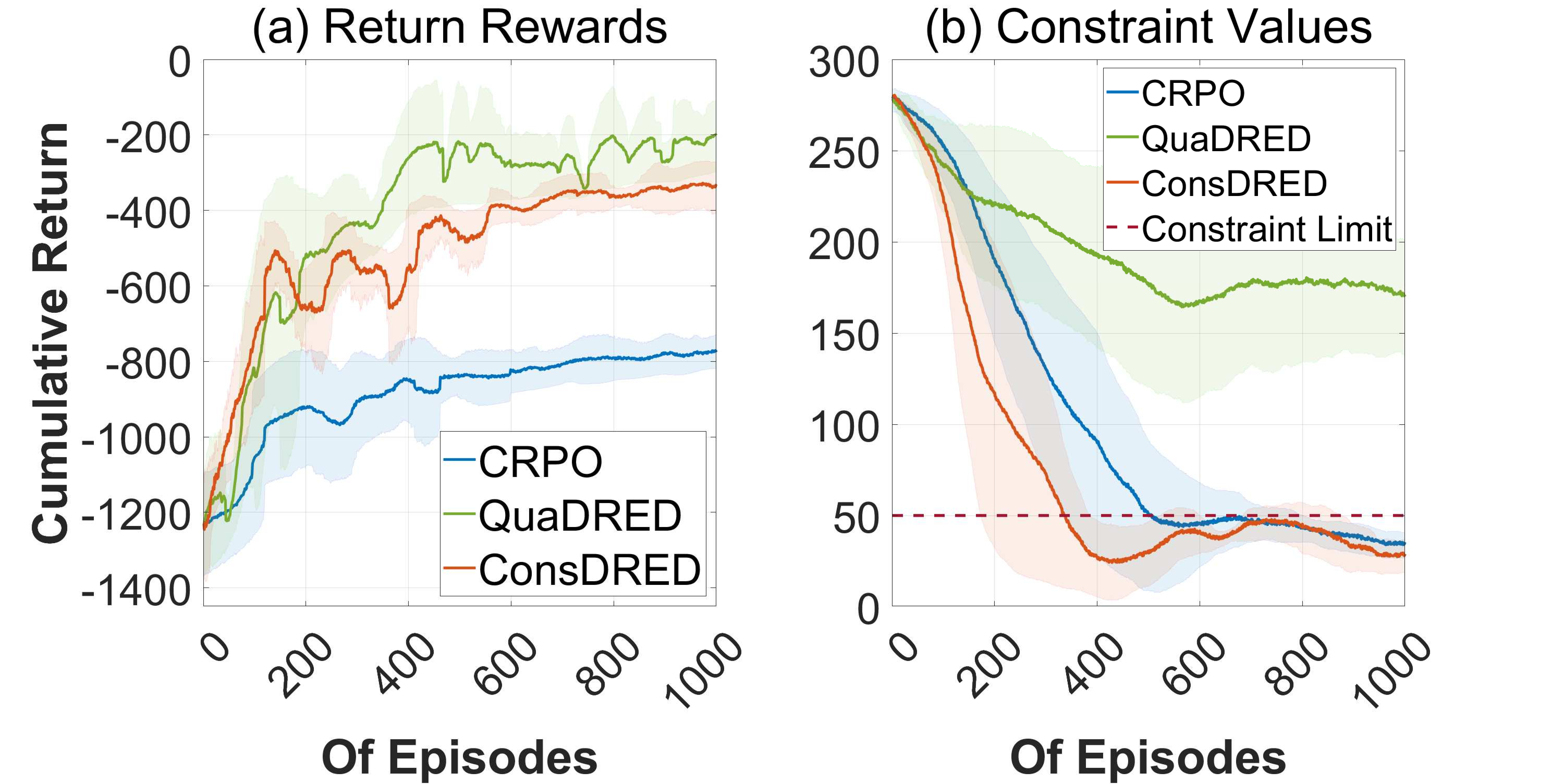}
  \caption{Learning curves of three RL algorithm: CRPO \cite{xu2021crpo}, QuaDRED \cite{wang2022interpretable} and ConsDRED. The simulated speed is set as $0.6$.}
  \label{learning_curves}
\end{figure}

Training directly in real physical environments is regularly performed by robotics researchers, given that such real-world training is known to avoid overfitting and helps to improve generalization, as necessary for shallow neural networks (e.g., the 2-hidden-layer networks used in our proposed ConsDRED). The whole training process is implemented firstly in the simulator before being redeployed in challenging real-world environments. We transition to the online physical training only when the corresponding constraint value converges below the constraint limit or demonstrates stable convergence if consistently above the limit, as depicted in Fig.~\ref{learning_curves} (b).

\noindent \textbf{Training Setting}: the quadrotor state $\bm{x}$ is recorded at 16 Hz. The whole training process occurs over 1000 iterations, within which the simulated training is terminated and transferred to the real-world training, described in Fig.~\ref{senarios_wind_no_obstacles}, Fig.~\ref{senarios_nowind_obstacles} and Fig.~\ref{senarios_wind_obstacles}.

The matrices $H_1$ and $H_2$ in \autoref{reward} are chosen as $H_1=diag\{ 2.5e^{-2}, 2.5e^{-2}, 2.5e^{-2}, 1e^{-3}, 1e^{-3}, 1e^{-3}, 2.5e^{-3}, \\ 2.5e^{-3}, 2.5e^{-3}, 2.5e^{-3}, 1e^{-5}, 1e^{-5}, 1e^{-5}\}$ and $H_2=diag\{ 1.25e^{-4}, 1.25e^{-4}, 1.25e^{-4}, 1.25e^{-4}\}$, respectively. Based on the benchmark \cite{wang2022kinojgm,zhang2019quota,xu2021crpo}, the parameters of our proposed framework are summarized in \autoref{model_parameters}. Note that we use `VehicleThrustSetpoint' and `VehicleTorqueSetpoint' as control inputs $\bm{u}$ to set the thrust and torque (FRD \footnote{FRD: coordinate system follows the right-hand rule, where the X-axis points toward the vehicle's Front, the Y-axis points Right, and the Z-axis points Down.}) in Pixhawk 4.

As shown in Algorithm~{\autoref{DRL_ConsDRED}}, the agent is rewarded at each time step unit, i.e., $1/16$ ($s$), by following \autoref{reward} whilst being penalized $+1$ at each time step unit, according to the constraints: (i) $(0,0,0)\leq \bm{P}_{WB,(X,Y,Z)}\leq (6.0,5.5,2.0)$ ($m$) and $0\leq {\left \| \bm{V}_{WB}\right \|}_2 \leq 12$ ($ms^{-1}$) in our experiments \footnote{These constraints are enforced due to the motor limits (\autoref{hardware_spec}) and the spatial confines of the experimental area. Recalling \autoref{quadrotor_dynamics}, both $\bm{P}_{WB}$ and $\bm{V}_{WB}$ are expressed in the world frame. ${\left \| \cdot\right \|}_2$ denotes the $l^2$-norm.}; and (ii) drifting into the unsafe areas shown in Fig.~\ref{simulation_env} under external disturbances.

\begin{figure}[t]
  \centering
  \includegraphics[scale=0.19]{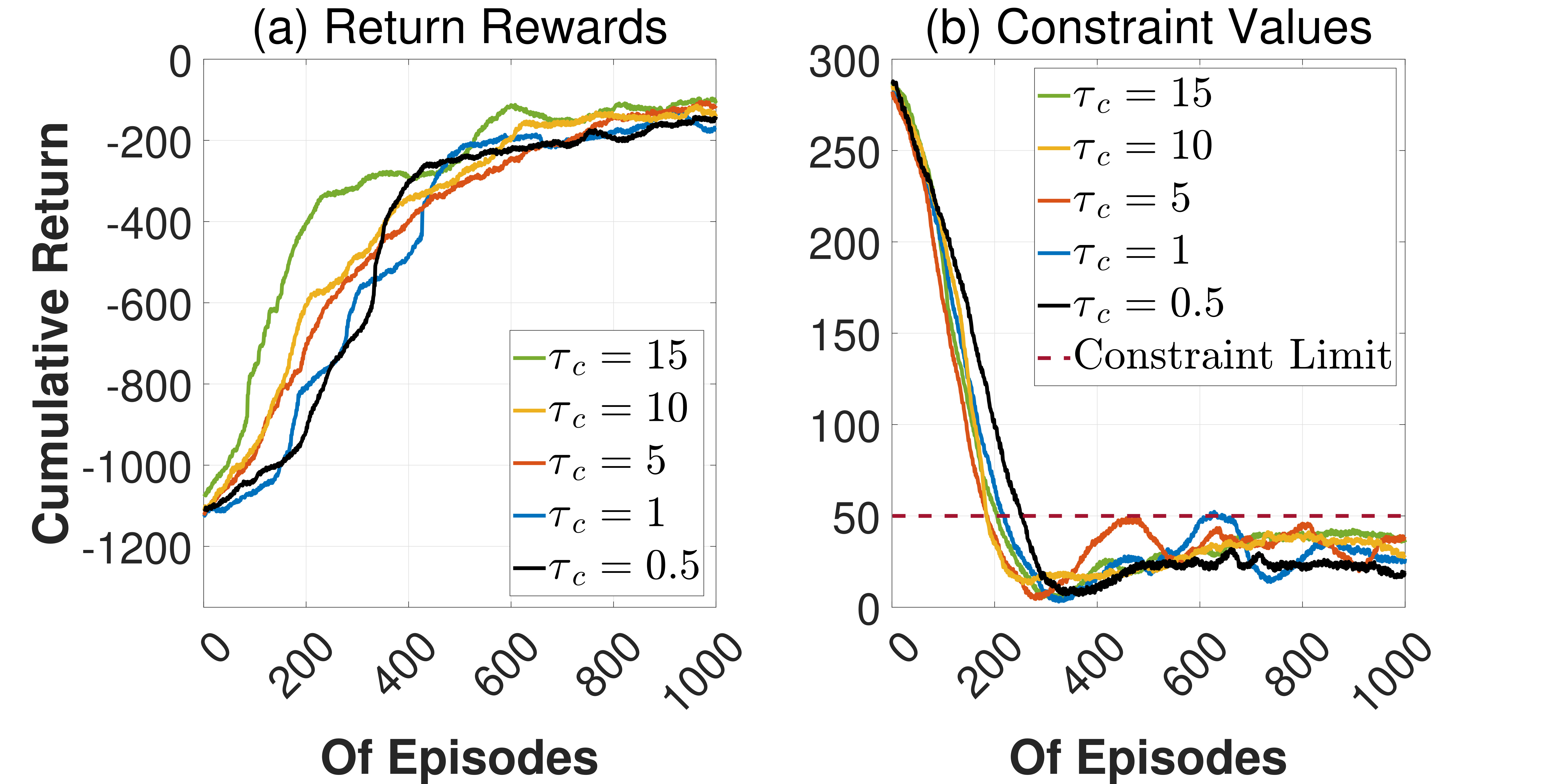}
  \caption{The robustness of different hyperparameters - i.e., the tolerance $\bm{\tau}_{c}$ setting - to the reward and constraint convergence process.}
  \label{tolerance}
\end{figure}

\noindent \textbf{Simulation Training}: our constrained simulation scenario is shown in Fig.~\ref{simulation_env}, where each episode has $T=200$. The reason for this setting is based on \textit{Theorem 2}: if let the episode length be $T=\Theta(\xi^{-2})=200$, then there exist $\overset{-}{CONV}_{T=200}=1-\frac{\mathcal{J}_r(\bm{\pi^*})-\mathbb{E}[\overset{-}{\mathcal{J}_r}(\bm{\pi}_{out})]}{\mathcal{J}_r(\bm{\pi^*})-\mathbb{E}[\overset{-}{\mathcal{J}_r}(\bm{\pi}_{0})]}\geq 1-\frac{\xi}{\xi_0}=1-\frac{\Theta(\frac{1}{(1-\gamma)\sqrt{T}})}{\Theta(\frac{1}{(1-\gamma)\sqrt{1}})}\approx93\%$. Thus we can guarantee at least $93\%$ convergence by setting the episode length $T=200$. The flight task in simulation is to track the seven reference waypoints. In the external disturbance zone, the quadrotor system operates with aerodynamic effects in the horizontal plane in the range [-3,3] ($ms^{-2}$).

\begin{figure*}[t]
  \centering
  \includegraphics[scale=0.42]{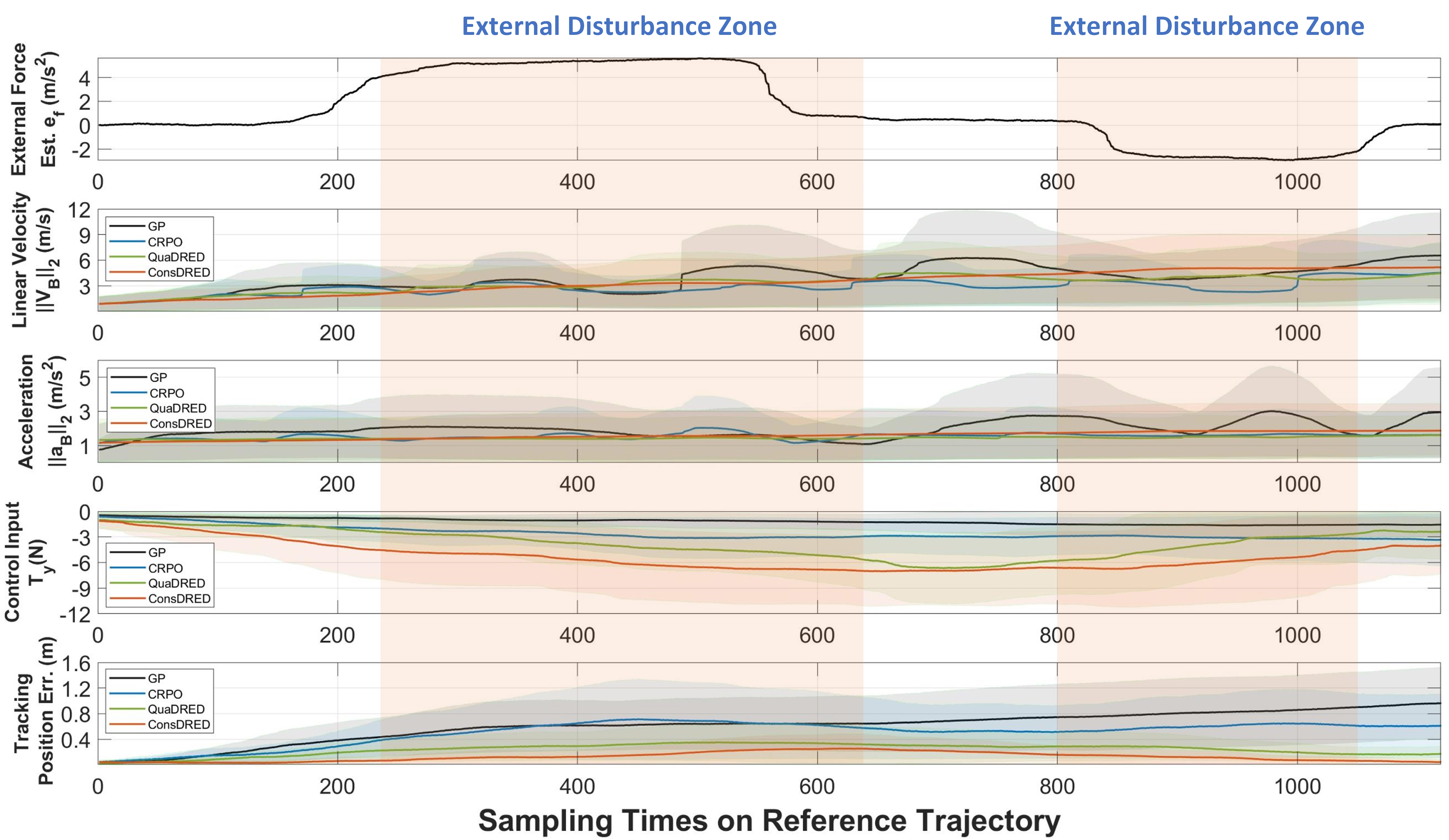}
  \caption{Simulation tracking performance assessed with variance measurement: the external force estimation $\bm{e}_{f}$($m/s^{2}$), linear velocity $\left \|\bm{V}_{B}\right \|_2$, acceleration $\left \|\bm{a}_{B}\right \|_2$, the control input $T_y$ (expressed in the body frame) and tracking position error ($m$). The X-axis unit corresponds to the number of sampling times, with 1100 sampling points evenly distributed along a trajectory. The results from a single run are depicted in Fig.~\ref{specific_scenarios_data}.}
  \label{simulated_results_variance}
\end{figure*}

\begin{figure*}[t]
  \centering
  \includegraphics[scale=0.455]{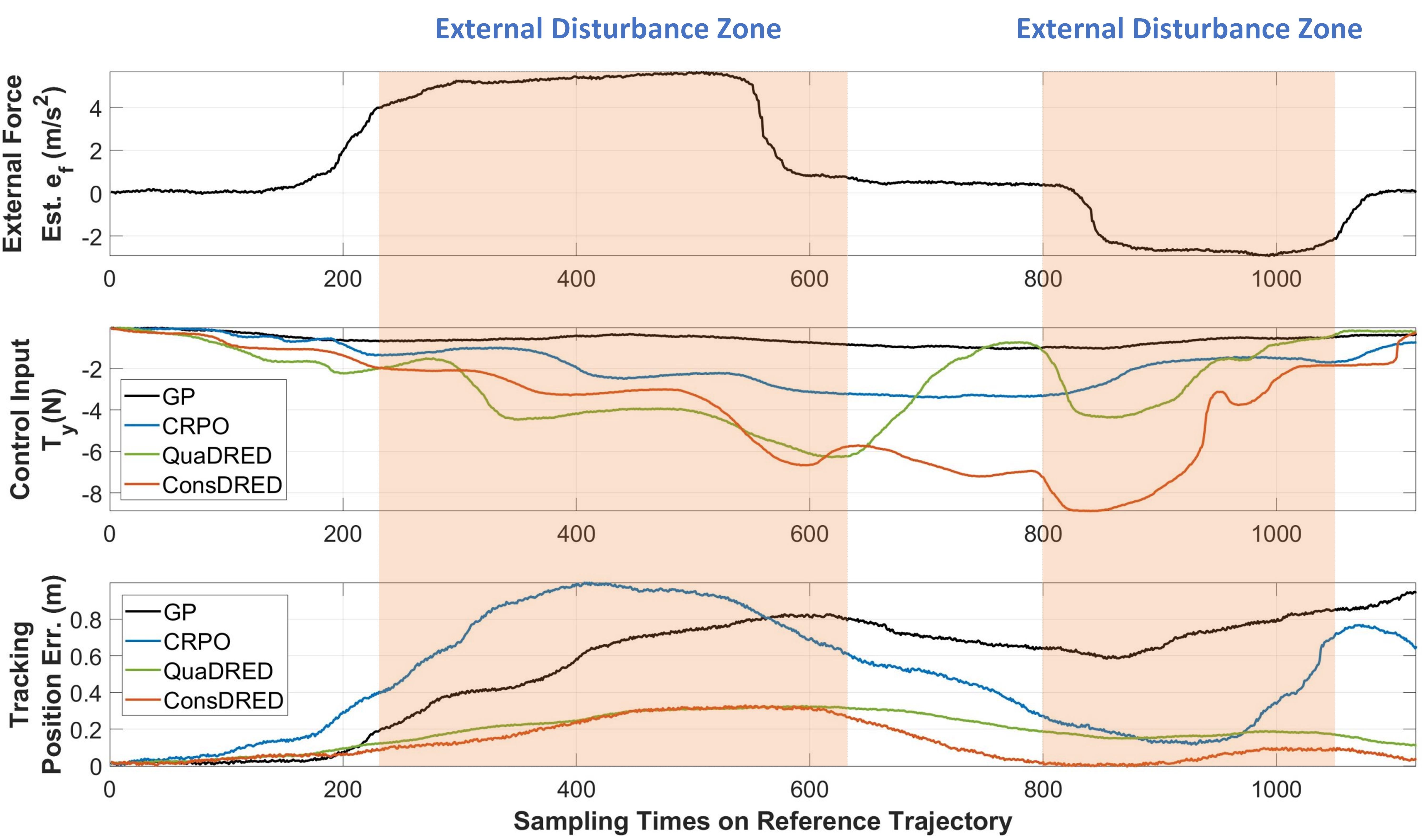}
  \caption{Simulation result from one run: the external force estimation $\bm{e}_{f}$, the control input $T_y$ (expressed in the body frame), and tracking position error $(m)$.}
  \label{specific_scenarios_data}
\end{figure*}

\noindent \textbf{Real-world Training}: Three scenarios are trained in the real physical world \footnote{Video figures in support are available: \url{https://github.com/Alex-yanranwang/ConsDRED-SMPC}.\label{video_footnote}}: trajectory tracking under external forces (i.e., winds generated from two fans) without obstacles (Flight Tasks 1-2 shown in Fig.~\ref{senarios_wind_no_obstacles}), trajectory tracking around the static (no external forces) but dense obstacles (Flight Tasks 3-4 shown in Fig.~\ref{senarios_nowind_obstacles}) and trajectory tracking under external forces around dense obstacles (Flight Task 5-6 shown in Fig.~\ref{senarios_wind_obstacles}). The magnitude of the external forces (i.e., winds) is in the range [0, 2.5] ($ms^{-1}$), where the winds are generated from the fans in Scenario 1 and 3.

The hardware specification can be found in \autoref{hardware_spec}. In our setup, we integrate a loop detection process into VID-Fusion \cite{ding2020vid}, following the methodology outlined in \cite{qin2018vins}, to enhance measurement accuracy. This leads to a significant reduction of $14.3\%$ in tracking error, improving the measurement accuracy {to} a range of [0.010, 0.021] ($m$).

\begin{table*}[t]
\caption{Simulated trajectory tracking under programmatic external forces, where the three external forces are: $\bm{E}_{f,1}=[0.0, 3.0, 0.0]$ ($ms^{-2}$), $\bm{E}_{f,2}=[-3.0, 3.0, 0.0]$ ($ms^{-2}$) and $\bm{E}_{f,2}=[-4.0, 4.0, 0.0]$ ($ms^{-2}$).}
\label{Comparison_of_sim_tracking}
\begin{center}
\setlength{\tabcolsep}{0.85mm}{
\begin{tabular}{c l c c c c c c c c c c c c}
\hline
\multirow{2}{*}{\textbf{External forces}}                        & \multirow{2}{*}{\textbf{Method}}      & \multirow{2}{*}{\textbf{\makecell{Success\\ times}}} & \multicolumn{2}{c}{\textbf{Time (s)}} & & \multicolumn{2}{c}{\textbf{Accu. error (m)}} & &\multicolumn{2}{c}{\textbf{RMSE (m)}} & & \multicolumn{2}{c}{\textbf{Cons. return}}\\ 
\cline{4-5} \cline{7-8} \cline{10-11} \cline{13-14}
& & & avg. & var.& & avg. & var. & & avg. & var. & & avg. & var. \\
\hline
\multirow{9}{*}{$\bm{E}_{f,1}$} & PD \cite{guo2022safety}        & 7/60              & 37.61      & 3.83& & 31.53  &5.13 & & 1.31  & 0.147 & &252 & 31.3                       \\ 
                                    & PD + VID-Fusion \cite{ding2020vid}        &  25/60             & 29.16      &3.48 & & 25.51  &4.20 & & 0.94  &0.123 & &239 & {25.4}                      \\ 
                                      & ADP \cite{dou2021robust} &   21/60               & 33.87      & {5.78}& & 29.43  &{6.17} & &0.95  &{0.175} & &246 & {41.9}                        \\ 
                                      & SMPC \cite{wang2022interpretable}        &  29/60               & 32.16      & {2.15}& & 25.48  &{4.85} & & 0.86  &{0.140} & &251 & {24.4}                   \\ 
                                      & SMPC + VID-Fusion &  47/60                & 17.31      &{1.43} & & 12.59  &{2.81} & &0.48  &{0.082} & &247 & {27.4}                        \\ 
                                        & GP \cite{torrente2021data}-SMPC       &  41/60              & 26.85      &{1.05} & & 21.62  &{4.16} & & 0.62  &{0.133} & &231 &{19.3}                      \\ 
                                      & CRPO \cite{xu2021crpo} + SMPC &  41/60                & 31.26      &{3.14} & & 19.58  &{3.78} & &0.58  &{0.115} & &36 & {8.8}                         \\ 
                                      & QuaDRED-SMPC \cite{wang2022interpretable}  &  55/60              & 10.58    & {1.85} & & 7.53  &{1.62} & & 0.21  &{0.042} & &141 & {12.5}                       \\
                                      & \bf{ConsDRED-SMPC}   & \bf{53/60}             & \bf{10.89}      &\bf{{1.12}} & & \bf{5.52}  &\bf{{1.55}} & & \bf{0.16}  &\bf{{0.052}} & &\bf{18} & \bf{{9.1}}
                                      \\ \hline
\multirow{9}{*}{$\bm{E}_{f,2}$} & PD        &  2/80             & -      &{-} & & -  &{-} & & -  &{-} & &-&{-}                       \\ 
                                    & PD + VID-Fusion         &  10/80             & 38.09      &{3.71} & & 29.28  &{4.77} & & 1.17  &{0.146} & &255 & {35.8}                       \\ 
                                      & ADP  &  15/80               & 41.16      &{5.94} & & 32.56  &{7.21} & &1.51  &{0.214} & &259& {40.5}                         \\ 
                                      & SMPC        &  9/80              & 40.67      &{3.83} & & 31.74  &{4.88} & & 1.33  &{0.145} & &242 &{28.4}                   \\ 
                                      & SMPC + VID-Fusion &  56/80               & 23.84      & {3.49}& & 13.83  &{3.11} & &0.50  &{0.094} & &246 &{23.5}                        \\ 
                                      & GP-SMPC         &  15/80              & 36.15     & {3.17}& & 24.11   &{4.02} & &0.87  &{0.135} & &273 &{20.7}                     \\ 
                                      & CRPO + SMPC &  50/80               & 28.33      &{4.37} & & 19.84  &{3.77} & & 0.67  &{0.117} & &31 &{10.1}                        \\ 
                                      & QuaDRED-SMPC &  66/80            & 12.20     &{3.35} & & 9.35  &{1.70} & &0.23  &{0.041} & &176 &{13.5}                        \\
                                      & \bf{ConsDRED-SMPC}    & \bf{67/80}             & \bf{14.11}      &\bf{{2.96}} & & \bf{9.89}  &\bf{{1.81}} & & \bf{0.26}  &\bf{{0.050}} & &\bf{33}&\bf{{11.2}}  
                                      \\ \hline
\multirow{9}{*}{$\bm{E}_{f,3}$} & PD        &  0/80             & -      &{-} & & -  &{-} & & -  &{-} & &-&{-}                       \\ 
                                    & PD + VID-Fusion         &  2/80              & -      &{-} & & -  &{-} & & -  &{-} & &- &{-}                      \\ 
                                      & ADP  & 8/80              & 42.26      &{6.26} & & 39.64  &{7.73} & &1.77  &{0.210} & &257 &{42.7}   \\ 
                                      & SMPC        &  2/80              & -      & {-}& & -  &{-} & &-  & {-}& &- &{-}                   \\ 
                                      & SMPC + VID-Fusion &  48/80               & 30.62      &{4.47} & & 24.07  &{4.08} & &0.84  &{0.125} & &250 &{27.1}                          \\ 
                                        & GP-SMPC         &  0/80             & -      &{-} & & -        & {-}& & - & {-}& &-&{-}                 \\ 
                                      & CRPO + SMPC &  41/80               & 28.56      &{5.63} & & 29.11  &{4.19} & & 0.97  &{0.116} & &35 &{14.5}                        \\ 
                                      & QuaDRED-SMPC &  67/80               & 13.62     &{3.28} & & 12.36   &{1.72} & & 0.34  &{0.052} & &189 &{16.2}                       \\
                                      & \bf{ConsDRED-SMPC}    & \bf{69/80}             & \bf{11.87}      &\bf{{3.45}} & & \bf{11.98}  &\bf{{1.83}} & & \bf{0.31}  &\bf{{0.058}} & &\bf{40}&\bf{{11.8}}  
                                      \\ \hline
\end{tabular}
}
\end{center}
\end{table*}

Each flight task is under surveillance from different perspectives, which are an external camera, {a} 3D occupancy grid map constructed based on \cite{zhou2021raptor}, and an on-board camera (i.e., Intel Realsense D435i). The measured trajectories are also shown from {the} top, front and left perspectives. Compared with the simulated training shown in Fig.~\ref{simulation_env}, the real-world training is more efficient for ConsDRED (i.e., the disturbance estimator) and can learn directly from the real aerodynamic effects, including a combination of the individual propellers and airframe \cite{hoffmann2007quadrotor}, turbulent effects caused by rotor–rotor and rotor-airframe interactions \cite{russell2016wind}, and other turbulences such as rotor-obstacle interactions \cite{kaya2014aerodynamic}.

\noindent \textbf{Convergence Performance in Training}: the learning curves are displayed in Fig.~\ref{learning_curves}, where we show the training performance of CRPO \cite{xu2021crpo}, QuaDRED \cite{wang2022interpretable} and our proposed ConsDRED. The performance shows that, in Fig.~\ref{learning_curves} (a), although all three algorithms converge to a long-term return eventually, the two distributional approaches, ConsDRED and QuaDRED, outperform the CRL approach, CRPO. QuaDRED achieves the best accumulated rewards. However, the unconstrained QuaDRED does not satisfy the constraint limit in Fig.~\ref{learning_curves} (b).

For the practical implementation of ConsDRED, we experiment with robustness with regard to the hyperparameter settings. The existing primal-dual approaches such as RCPO \cite{tessler2018reward}, CPPO \cite{stooke2020responsive}, and PPO \cite{ding2021provably} are very sensitive to the hyperparameters. Thus{,} additional costs incurred for hyperparameter tuning are inefficient for practical implementation and can hinder important applications, including real-world robots with high uncertainties. Setting the tolerance $\bm{\tau}_{c}$ variably as $\left \{0.5,1,5,10,15\right \}$, Fig.~\ref{tolerance} demonstrates that ConsDRED's convergence is robust with respect to the tolerance $\bm{\tau}_{c}$ over the whole training process. Therefore, we can conclude that the tolerance $\bm{\tau}_{c}$ will not incur large tuning cost when ConDRED is being implemented.

\subsection{Comparative performance of ConsDRED-SMPC under variable aerodynamic effects}
\noindent \textbf{Simulated Tracking Performance under Programmatic External Forces}: we compare our ConsDRED-SMPC against a state-of-the-art trajectory tracking algorithm, GP\footnote{Followed by the baseline \cite{torrente2021data}, we gathered a dataset comprising velocities [-12,12] ($ms^{-1}$) to train the GP. This dataset is obtained by tracking randomly generated aggressive trajectories, as shown in Fig.~\ref{simulation_env}.} \cite{torrente2021data}-SMPC, and interactive approaches, CRPO \cite{xu2021crpo} and QuaDRED \cite{wang2022interpretable}, with variable aerodynamic forces added to our simulated environment to evaluate the over-fitting problems of RL. The experiments are based on the trained ConsDRED model described in Section \uppercase\expandafter{\romannumeral6}-A. We first set the aerodynamic forces as [0.0, 3.0, 0.0] ($ms^{-2}$). In Fig.~\ref{simulated_results_variance} and Fig.~\ref{specific_scenarios_data}, the tracking position errors and control inputs are depicted for two opposite heading aerodynamic forces (both having the same force [0.0, 3.0, 0.0] ($ms^{-2}$)). Notably, our proposed ConsDRED-SMPC demonstrates the smallest tracking position error and exhibits an efficient response to sudden aerodynamic effects in these specific scenarios.

Then two larger and more complex forces, i.e., [-3.0, 3.0, 0.0] and [-4.0, 4.0, 0.0] ($ms^{-2}$), are used in the scenario shown in Fig.~\ref{simulation_env}.  To achieve more accurate outcomes, the number of experiments in \autoref{Comparison_of_sim_tracking} is elevated from 60 to 80 in the second and third instances. To comprehensively assess the performance, we conduct a thorough comparison in \autoref{Comparison_of_sim_tracking}, involving various controllers: `PD \cite{guo2022safety}' (a standard controller \footnote{We tune parameters by initially employing manual tuning and subsequently performing auto-tuning, as outlined in \cite{nguyen2018development}, within the PX4 user guide \cite{meier2015px4}.}), `PD + VID-Fusion \cite{ding2020vid}' (a standard adaptive controller), `ADP \cite{dou2021robust}' (a combination of dynamic programming and RL \footnote{The parameter setting is derived from \cite{dou2021robust}, wherein the optimal parameters are dynamically learned online based on the quadrotor state.}), `SMPC' (a baseline controller), `SMPC + VID-Fusion' (a combination of a baseline controller and an adaptive estimator), `GP-SMPC' (a combination of GP \cite{torrente2021data} and SMPC), `CRPO + SMPC' (a combination of CRPO \cite{xu2021crpo} and SMPC), `QuaDRED-SMPC' (a combination of QuaDRED \cite{wang2022interpretable}, DDPG \cite{lillicrap2015continuous}, and SMPC), and `ConsDRED-SMPC' (our proposed method). {Analyzing the outcomes in TABLE \uppercase\expandafter{\romannumeral4}, it becomes evident that the adaptive controller 'PD + VID-Fusion' performs poorly. This is primarily attributed to the inherent limitations of the vision-inertial measurement module in VID-Fusion \cite{ding2020vid}, leading to inaccurate estimations, especially in larger aerodynamics. To address this issue, a disturbance estimator such as ConsDRED and QuaDRED \cite{wang2022interpretable} is introduced into the control framework, as depicted in Fig.~\ref{RL_Control_framework}. Further insights from the results in TABLE \uppercase\expandafter{\romannumeral4} reveal that `PD' and `PD + VID-Fusion' exhibit poor performance. Consequently, we exclude these for real-world experiments shown in TABLE \uppercase\expandafter{\romannumeral5}.

\begin{table}[t]
    \caption{Technical Specification of Hardware}
    \label{hardware_spec}
    \setlength{\tabcolsep}{0.8mm}{
    \begin{tabular}{c c c}
    \toprule
    \textbf{No.} & \textbf{Component} & \textbf{Specific Model} \\
    \hline
    1 & Frame & QAV250 \\
    2 & Sensor - Depth Camera & Intel Realsense D435i \\
    3 & Sensor - Down-view Rangefinder & Holybro ST VL53L1X \\
    4 & Flight Controller & Pixhawk 4 \\
    5 & Motors & T-Motor F60 Pro IV 1750KV \\
    6 & Electronic Speed Controller & BLHeli-32bit 45A 3-6s \\
    7 & On-board Companion Computer & \makecell[c]{DJI Manifold 2-c\\ (CPU Model: Intel Core i7-8550U)}  \\
    8 & Mounts & \makecell[c]{3D Print for Sensors/\\Computer/Controller/Battery} \\
    \bottomrule
    \end{tabular}}
\end{table}

\begin{figure}[t]
  \centering
  \includegraphics[scale=0.105]{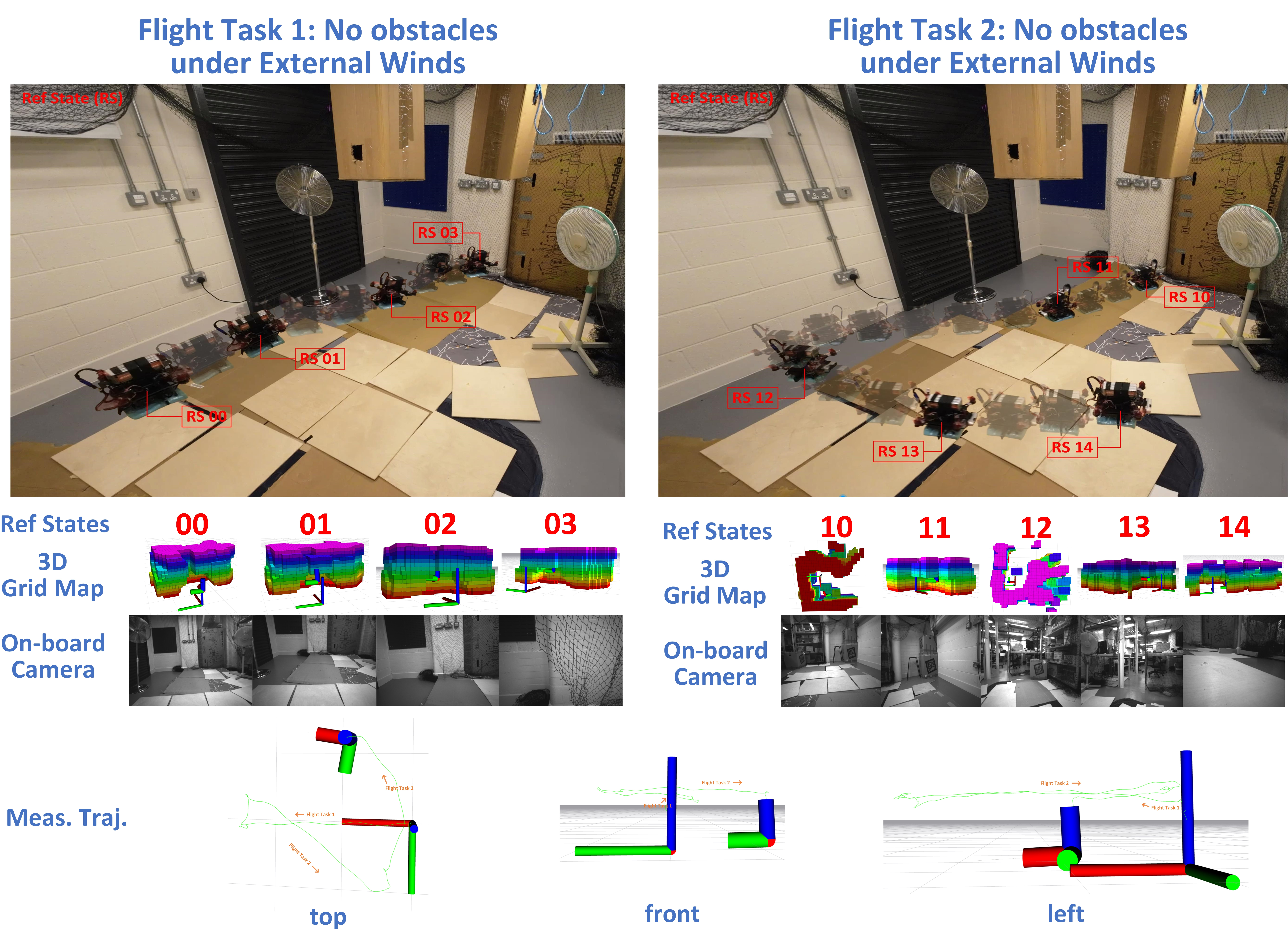}
  \caption{Real-world Scenario 1: tracking trajectories under external forces without obstacles, where the agent learns the aerodynamic effects like turbulent effects caused by dynamic rotor–rotor and rotor-airframe interactions \cite{hoffmann2007quadrotor} in Flight Task 1-2.}
  \label{senarios_wind_no_obstacles}
\end{figure}

\begin{figure}[t]
  \centering
  \includegraphics[scale=0.105]{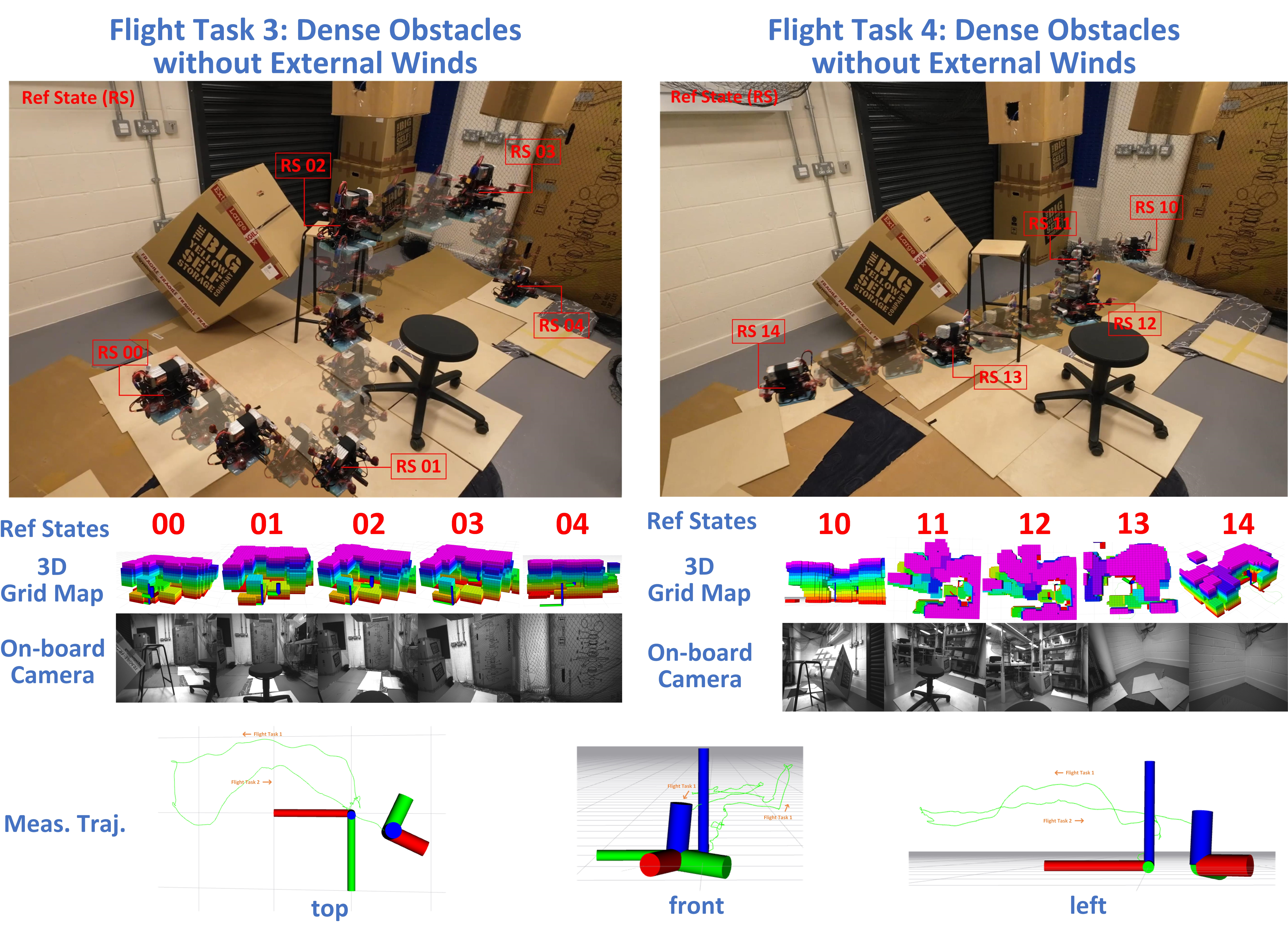}
  \caption{Real-world Scenario 2: tracking trajectories around the static (no external forces) but dense obstacles, where the agent learns the turbulent effects from rotor-obstacle interactions \cite{kaya2014aerodynamic} in Flight Task 3-4.}
  \label{senarios_nowind_obstacles}
\end{figure}

\begin{figure}[t]
  \centering
  \includegraphics[scale=0.105]{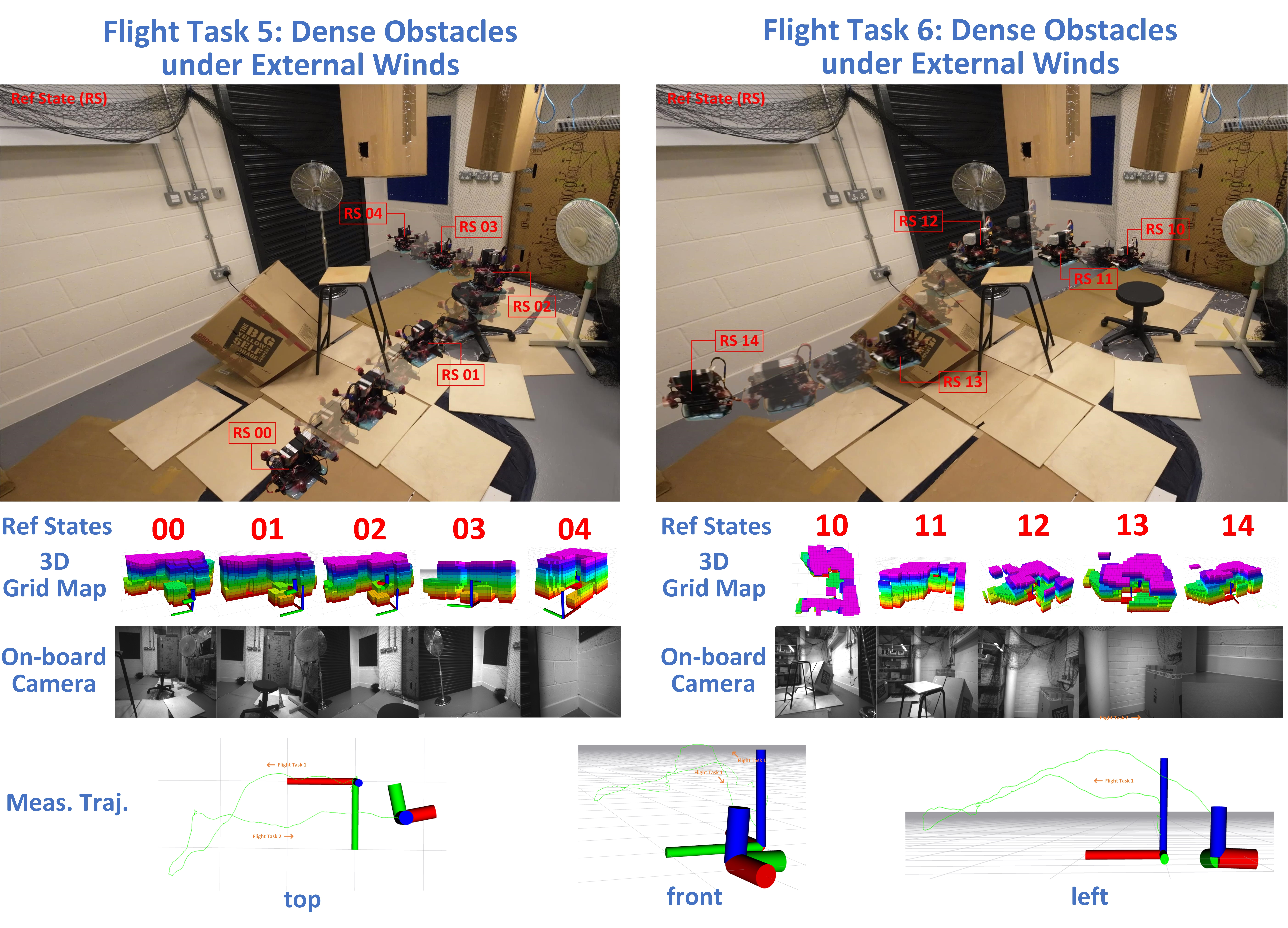}
  \caption{Real-world Scenario 3: tracking trajectories under external forces around dense obstacles, where the agent learns the comprehensive aerodynamic effects from a combination of the dynamic rotor–rotor, rotor-airframe, and rotor-obstacle interactions in Flight Task 5-6.}
  \label{senarios_wind_obstacles}
\end{figure}

Our results also show that interactive approaches are not always greater than non-interactive approaches. For example, `CRPO + SMPC' has a lower success\footnote{A successful flight: the quadrotor completes the flight from the starting point to the destination point, allowing for minor collisions without resulting in a collapse.} times than GP-SMPC, whilst there is little difference in the operation time and accumulative tracking error with relatively small aerodynamic forces. However, compared with GP-SMPC, our proposed ConsDRED-SMPC achieves improvements of $>61\%$ in operation time, $>70\%$ in RMSE \footnote{The RMSE metric in both \autoref{Comparison_of_sim_tracking} and \autoref{Comparison_of_real_tracking} represents the average RMSE over all successful trajectories for a specific controller.} tracking errors, and $>87\%$ in accumulative constrained returns, respectively. Intuitively, the superior performance of both distributional RL and SMPC over their conventional counterparts stems from their more comprehensive and enriched representations — i.e., the entire value distribution in distributional RL and probabilistic descriptions in SMPC. These approaches help avoid the worst-case conservatism. The two distributional RL approaches, i.e., `QuaDRED-SMPC' and `ConsDRED-SMPC', have closed success times, operation time and tracking errors, where `QuaDRED-SMPC' has better performance in some cases. However, compared to `QuaDRED-SMPC', our `ConsDRED-SMPC' achieves $87.2\%$, $81.3\%$ and $78.8\%$ improvements in constrained returns.

The simulated results highlight a significant portion of the performance improvement arising from the integration of a wind estimator (VID-Fusion). This integration leads to a $23\%$ improvement in relatively small wind conditions and a substantial $46\%$ improvement in more extensive wind conditions. Moreover, the trend indicates that the contribution of ConsDRED to performance becomes more pronounced with increasing external wind intensity, escalating from $54\%$ to $77\%$.

\noindent \textbf{Real-world Tracking Performance under Variable Disturbances}: The aim is to evaluate two properties of the trained ConsDRED: 1) convergence quality in the real physical experiments, and 2) generalization capability under unprecedented external forces.

\begin{table*}[t]
\caption{Real-world trajectory tracking under variable external forces: the scenario 1 (shown in Fig.~\ref{senarios_wind_no_obstacles}) and 3 (shown in Fig.~\ref{senarios_wind_obstacles}) are applied for tracking performance evaluation. Let the external forces from the left / right fans in scenario 1 / 3 be $\bm{F1}_{l,s3}=2.5$ ($ms^{-1}$), $\bm{F1}_{r,s3}=1.5$ ($ms^{-1}$); $\bm{F2}_{l,s1}=3.5$ ($ms^{-1}$), $\bm{F2}_{r,s1}=2.5$ ($ms^{-1}$); and $\bm{F3}_{l,s3}=3.5$ ($ms^{-1}$), $\bm{F3}_{r,s3}=2.5$ ($ms^{-1}$).}
\label{Comparison_of_real_tracking}
\begin{center}
\setlength{\tabcolsep}{0.92mm}{
\begin{tabular}{c l c c c c c c c c c c c c}
\hline
\multirow{2}{*}{\textbf{External forces}}                        & \multirow{2}{*}{\textbf{Method}}      & \multirow{2}{*}{\textbf{\makecell{Success\\ times}}} & \multicolumn{2}{c}{\textbf{Time (s)}} & & \multicolumn{2}{c}{\textbf{Accu. error (m)}} & &\multicolumn{2}{c}{\textbf{RMSE (m)}} & & \multicolumn{2}{c}{\textbf{Cons. return}}\\ 
\cline{4-5} \cline{7-8} \cline{10-11} \cline{13-14}
& & & avg. & var.& & avg. & var. & & avg. & var. & & avg. & var. \\
\hline
\multirow{6}{*}{\makecell[l]{$\bm{F1}_{l,s3}$\\$\bm{F1}_{r,s3}$}}  &SMPC     & 12/20            & 17.76   &{1.40} & & 4.12 &{0.76} & & 0.314 &{0.051} & & 239 & {37.8}                   \\ 
                                      &  SMPC + VID-Fusion & 17/20               & 16.02   &{1.61} & & 1.95 &{0.82} & &0.120 &{0.044} & &224 & {28.6}\\ 
                                      & GP-SMPC      & 16/20             & 15.95    &{0.96} & &  3.81 &{0.57} & &  0.303 &{0.036}& &  216 & {29.3}                     \\ 
                                      & CRPO + SMPC & 14/20               & 17.09    &{1.32} & & 2.10 &{0.45} & & 0.119 &{0.037} & &30 & {7.6}                      \\ 
                                      & QuaDRED-SMPC & 18/20              & 15.56    &{0.58} & & 1.85 &{0.28} & & 0.103 &{0.021} & &111 & {17.3}                      \\
                                      & \bf{ConsDRED-SMPC}    & \bf{18/20}             & \bf{15.91}     &\bf{{0.53}} & & \bf{1.91} &\bf{{0.31}} & & \bf{0.106} &\bf{{0.026}} & & \bf{33}& \bf{{8.5}}
                                      \\ \hline
\multirow{6}{*}{\makecell[l]{$\bm{F2}_{l,s1}$\\$\bm{F2}_{r,s1}$}} &SMPC        & 2/20             & 18.52     &{0.90} & & 5.13 &{0.62} & &0.572 &{0.068} & &249 & {28.0}                 \\ 
                                      & SMPC + VID-Fusion & 13/20               & 17.67     &{1.28} & & 2.68 &{0.79} & &0.301 &{0.037} & &245& {32.6}\\ 
                                      & GP-SMPC       & 7/20              & 17.20    &{1.01} & & 3.58 &{0.61} & & 0.302 &{0.030} & & 210 &{27.1}                     \\ 
                                      & CRPO + SMPC & 13/20              & 17.41    &{1.35} & & 2.08 &{0.63} & & 0.151 &{0.032} & & 32  & {7.8}                       \\ 
                                      & QuaDRED-SMPC  & 20/20            & 16.35   &{0.70} & & 1.65 &{0.23} & & 0.102 &{0.019} & &97  & {16.7}                     \\
                                      & \bf{ConsDRED-SMPC}   & \bf{20/20}             & \bf{16.52}     &\bf{{0.57}} & & \bf{1.71} &\bf{{0.38}} & & \bf{0.102} &{0.022} & & \bf{18} &\bf{{6.2}}
                                      \\ \hline
\multirow{6}{*}{\makecell[l]{$\bm{F3}_{l,s3}$\\$\bm{F3}_{l,s3}$}}&SMPC        & 0/30              & -     & {-} & & - &{-} & &- &{-} & &- & {-}                   \\ 
                                      & SMPC + VID-Fusion & 7/30               & 18.03     &{1.31} & & 3.71 &{0.83} & &0.296 &{0.052} & &237& {33.6}\\ 
                                      & GP-SMPC         & 2/30              & 18.52     &{1.08} & & 4.83     &{0.66} & & 0.534 &{0.057} & & 238 & {25.0}                \\ 
                                      & CRPO + SMPC & 6/30               & 17.52     &{1.43} & & 3.52 &{0.84} & & 0.276 &{0.033} & &41 & {8.8}                      \\ 
                                      & QuaDRED-SMPC & 25/30               & 16.12    &{0.97} & & 2.82  &{0.35} & &0.192 &{0.024} & &156 & {21.6}                      \\
                                      & \bf{ConsDRED-SMPC}    & \bf{28/30}             & \bf{15.98}     &\bf{{0.62}} & & \bf{2.06} &\bf{{0.34}} & & \bf{0.109} &{0.016} & &\bf{41}& \bf{{10.7}}
                                      \\ \hline
\end{tabular}
}
\end{center}
\end{table*}

We first evaluate the convergence quality in Scenario 3 (shown in Fig.~\ref{senarios_wind_obstacles}). The external forces from the left / right fans are represented as $\left \{\bm{F1}_{l,s3},\bm{F1}_{r,s3}\right \}$, both of which are within training range, [0, 2.5] ($ms^{-1}$) generated from fans. We only change the obstacle positions slightly and keep the same number and similar density of obstacles. The quadrotor is required to perform the flight task described in Fig.~\ref{senarios_wind_obstacles}, where the reference trajectories are generated by Kino-JSS \cite{wang2022kinojgm}. The generalization capability is next evaluated in Scenario 1 and 3, where the feeding external forces are unprecedented, i.e., out of the range [0, 2.5] ($ms^{-1}$). $\left \{ \bm{F2}_{l,s1},\bm{F2}_{r,s1}\right \}$ and $\left \{\bm{F3}_{l,s3},\bm{F3}_{r,s3}\right \}$ denotes the external forces from the left / right fans shown in Fig.~\ref{senarios_wind_no_obstacles} and Fig.~\ref{senarios_wind_obstacles}, respectively.

The first case in \autoref{Comparison_of_real_tracking} shows again in real flight tasks that compared with `SMPC + VID-Fusion', RL-based approaches are not always significantly better, especially under relatively small external disturbances, which verifies empirically the conclusion in \cite{torrente2021data,wang2022interpretable}. In terms of the convergence quality, although all approaches can complete the tasks in most instances, the two distributional RL approaches significantly outperform `SMPC', `GP\footnote{To fit the GPs, flight data from the real world is collected, as detailed in Section \uppercase\expandafter{\romannumeral3}-F in \cite{torrente2021data}, and visualized in the accompanying video \footref{video_footnote}.}-SMPC' and `CRPO + SMPC'. Then, the second and third cases show good generalization ability of the proposed `ConsDRED-SMPC', which achieves comprehensively improved results in operation time, accumulative tracking errors, and constrained returns. Since `GP-SMPC' and `CRPO + SMPC' have low `Success Times' ($<4$ in the first $20$ experiments), to obtain a more accurate value, the experimental number in \autoref{Comparison_of_real_tracking} increases from 20 to 30 in the third case.

Compared with the two interactive RL approaches, i.e., constrained `CRPO + SMPC' and the unconstrained `QuaDRED-SMPC', we also find that ConsDRED-SMPC balances a trade-off between pursuing higher performance and obeying safety constraints: (i) `QuaDRED' outperforms our proposed `ConsDRED' in some cases, but it does not consider the constraints for safety taking an unconstrained approach; and ii) `CRPO' behaves as safely as `ConsDRED' (`Cons. Return' in \autoref{Comparison_of_real_tracking}), however, it performs poorly on the success times, operation time and tracking errors, because of its conservative decisions.

The influence of ConsDRED becomes even more evident in this real-world tracking experiments, shown in TABLE \uppercase\expandafter{\romannumeral5}. The contribution of ConsDRED rises significantly from $6.7\%$ to $57\%$ as the wind intensifies, transitioning from $\bm{F1}_{l,s3}=2.5$ ($ms^{-1}$), $\bm{F1}_{r,s3}=1.5$ ($ms^{-1}$) to $\bm{F2}_{l,s1}=3.5$ ($ms^{-1}$), $\bm{F2}_{r,s1}=2.5$ ($ms^{-1}$).

\section{Conclusion}
We propose ConsDRED-SMPC, an accurate trajectory tracking framework for quadrotors operating in environments with variable aerodynamic forces. ConsDRED-SMPC combines aerodynamic disturbance estimation and stochastic optimal control to address the aerodynamic effects on quadrotor tracking. A constrained distributional RL with quantile approximation, ConsDRED, is developed to improve the accuracy of aerodynamic effect estimation, where it achieves an $\Theta(1/{\sqrt{T}})$ convergence rate to the global optimum whilst an $\Theta(1/{m^{\frac{H}{4}}})$ approximation error. Using SADF for control parameterization to guarantee convexity, {an} SMPC is used to avoid conservative control returns and significantly improves the accuracy of quadrotor tracking. The aerodynamic disturbances are considered to have non-zero mean in the entire ConsDRED-SMPC framework. For practical implementation, we empirically demonstrate ConsDRED's convergence for training in simulation before moving to real-world training and validation. In contrast to most existing CRL approaches, our proposed ConsDRED is less sensitive to the tolerance $\bm{\tau}_{c}$ setting, i.e., hyperparameter tuning, and is hence easier to tune. Finally, we demonstrate that, in both simulated and real-world experiments, our proposed approach can track aggressive trajectories accurately under complex aerodynamic effects while guaranteeing both the convergence of ConsDRED and the stability of the whole control framework.

Our ongoing work concerns demonstrating the concept of recursive feasibility in SMPC, and future work will involve additional data collection and training processes in the wind tunnel to precisely identify the turbulent effects. We also prioritize investigating outdoor environments with additional unknowns and complexities. To reduce the computational complexity, we will implement ConsDRED-SMPC on dedicated hardware, for example, via FPGA implementation.

\bibliographystyle{unsrt}
\bibliography{ref}

\begin{IEEEbiography}[{\includegraphics[width=1in,height=1.25in,clip,keepaspectratio]{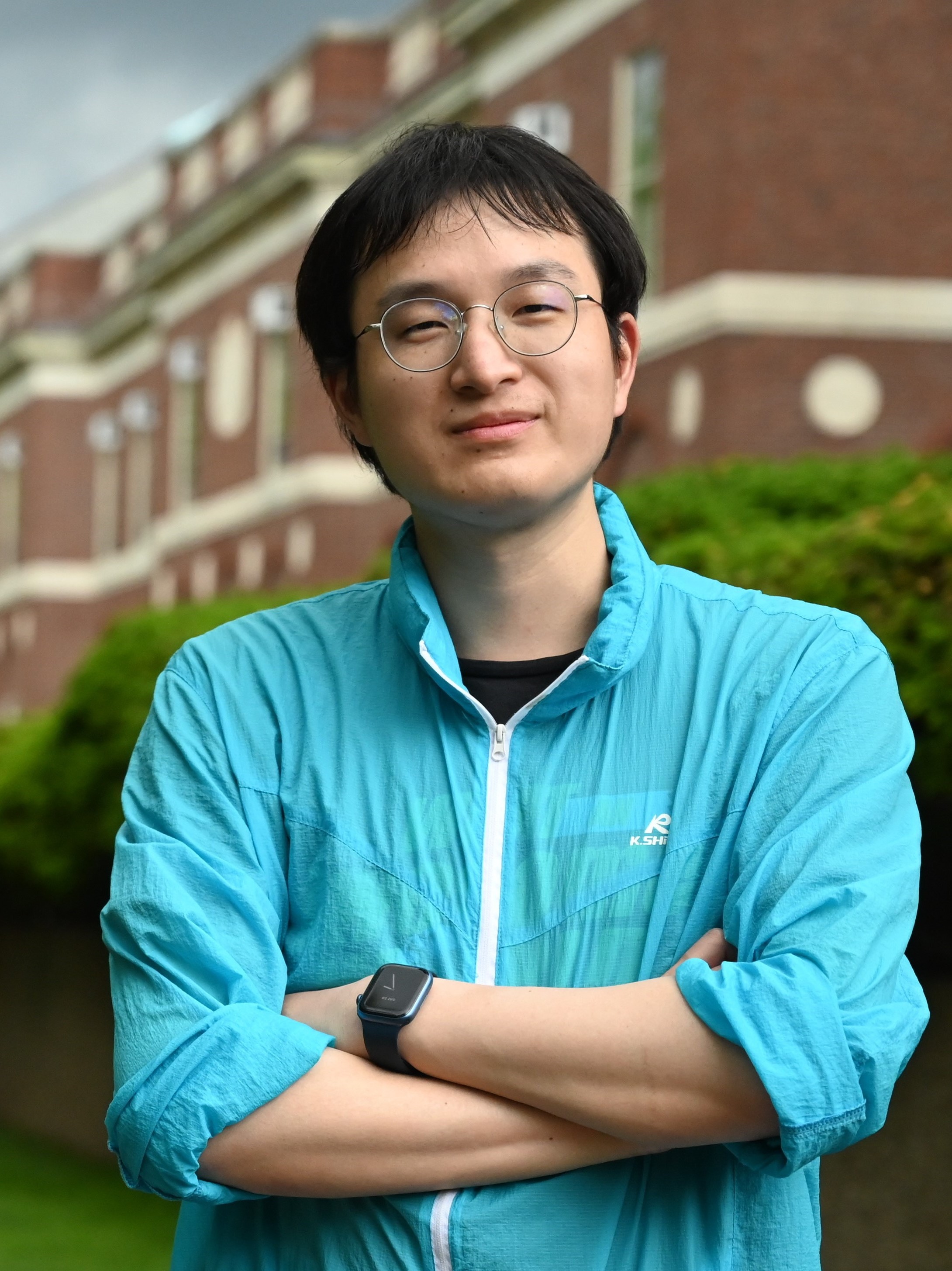}}]{Yanran Wang} received the M.Sc. degree in Aeronautics and Astronautics Science and Technology at Shanghai Jiao Tong University, Shanghai, China, in 2020, and B.Sc. degree in Automation at Southeast University, Nanjing, China, in 2017. He is currently pursuing a Ph.D. degree at Imperial College London. His research interests mainly include reliable learning-based controllers in cyber-physical Systems.
\end{IEEEbiography}

\begin{IEEEbiography}[{\includegraphics[width=1in,height=1.25in,clip,keepaspectratio]{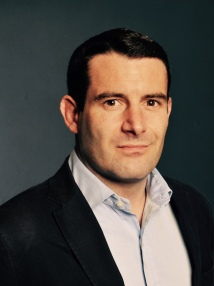}}]{David Boyle} (Member, IEEE) is an Associate Professor (Senior Lecturer) with the Dyson School of Design Engineering, Imperial College London. He received his B.Eng. and Ph.D. degrees in Computer and Computer and Electronic Engineering from the University of Limerick, Ireland, in 2005 and 2009, respectively. His research interests include the design of secure, private, and trustworthy cyber-physical systems.
\end{IEEEbiography}

\end{document}